\documentclass[lettersize,journal]{IEEEtran}
\usepackage{amsmath,amsfonts}
\usepackage{algorithmic}
\usepackage{algorithm}
\usepackage{array}
\usepackage[caption=false,font=normalsize,labelfont=sf,textfont=sf]{subfig}
\usepackage{textcomp}
\usepackage{stfloats}
\usepackage{url}
\usepackage{verbatim}
\usepackage{graphicx}
\usepackage{cite}
\usepackage{amssymb}
\usepackage{mathtools}
\usepackage{amsthm}

\usepackage{multirow}
\usepackage{enumitem}
\usepackage{booktabs}
\usepackage{hyperref}

\begin{document}

\title{CoLink: Global Compression and Knowledge Linking for Multi-Turn Dialogue Modeling}

\author{Lian Yan, Zhenbo Xia, Haotian Wang, Chen Tang, Yi Guan, Jie Liu,~\IEEEmembership{Fellow,~IEEE,} Jingchi Jiang
\thanks{* Corresponding author: Jingchi Jiang}
\thanks{Lian Yan, Zhenbo Xia, Haotian Wang, and Yi Guan are with the Research Center of Language Technology, Faculty of Computing, Harbin Institute of Technology, Harbin, Heilongjiang 150001, China (e-mail: yanlian0216@163.com; 23S103141@stu.hit.edu.cn; wanght1998@gmail.com; guanyi@hit.edu.cn).}
\thanks{Jie Liu and Jingchi Jiang are with the Research Center of Artificial Intelligence of Things, Faculty of Computing, Harbin Institute of Technology, Harbin, Heilongjiang 150001, China, and also with the National Key Laboratory ofSmart Farm Technologies and Systems, Harbin, Heilongjiang 150001, China (e-mail: jieliu@hit.edu.cn; jiangjingchi@hit.edu.cn).}
\thanks{Chen Tang is with Institute for Advanced Algorithms Research, Shanghai, 201306, China (e-mail: travistang@foxmail.com).}
\thanks{The code is available at \protect\url{https://github.com/YanPioneer/CoLink}.}
\vspace{-6mm}
}


\maketitle

\begin{abstract}
Extracting salient information from long-horizon dialogue histories remains a significant challenge for Large Language Models (LLMs) due to the restricted context windows and efficient retrieval challenges. Current methods typically rely on history organization or local token-level optimization, suffering from information loss, retrieval redundancy, and myopic attention. Inspired by the view in cognitive science that dialogue constitutes sense-making and memory association, we propose CoLink, a framework that enhances LLMs' global context awareness in domain-specific multi-turn dialogues through information linking and compression. CoLink first clusters historical contexts into topic graphs to establish cross-semantic associations with external knowledge. Guided by the adaptive control of thought, it then retrieves and activates knowledge via synergizing semantic relevance and graph-structural topology. Crucially, to achieve effective compression of history and knowledge, we ground multi-turn dialogues in information theory and derive a principled optimization objective for extracting the core sense-making information. Coupled with a competitive slot mechanism, CoLink condenses the global context into dedicated memory and knowledge slots, efficiently distilling lengthy, redundant inputs into an extremely compact latent representation. Extensive empirical results on both public and our MMD datasets validate CoLink’s superiority and generalizability in knowledge linking and sense modeling.
\end{abstract}

\begin{IEEEkeywords}
Multi-turn dialogue, global information compression, knowledge adaptive activation, information bottleneck.
\end{IEEEkeywords}

\section{Introduction}\label{sec:introduction}
\IEEEPARstart{L}{arge} Language Models (LLMs) effectively capture real-world knowledge and dialogue patterns via the long-range dependency modeling of Transformer, achieving remarkable performance in summarization, comprehension, and multi-hop reasoning \cite{achiam2023gpt,jiang2025knowledge,qiang2026exploring}. However, scaling these capabilities to long-context scenarios remains a formidable challenge. The quadratic complexity of self-attention and restricted context windows constrain full historical perception, while the accumulation of redundant information often obscures critical semantic associations \cite{li2024streamingdialogue,chen2024core}. Consequently, efficiently extracting contextually relevant memory and knowledge from histories is crucial for advancing multi-turn understanding and generation.

\begin{figure}[]
    \centerline{\includegraphics[width=\columnwidth]{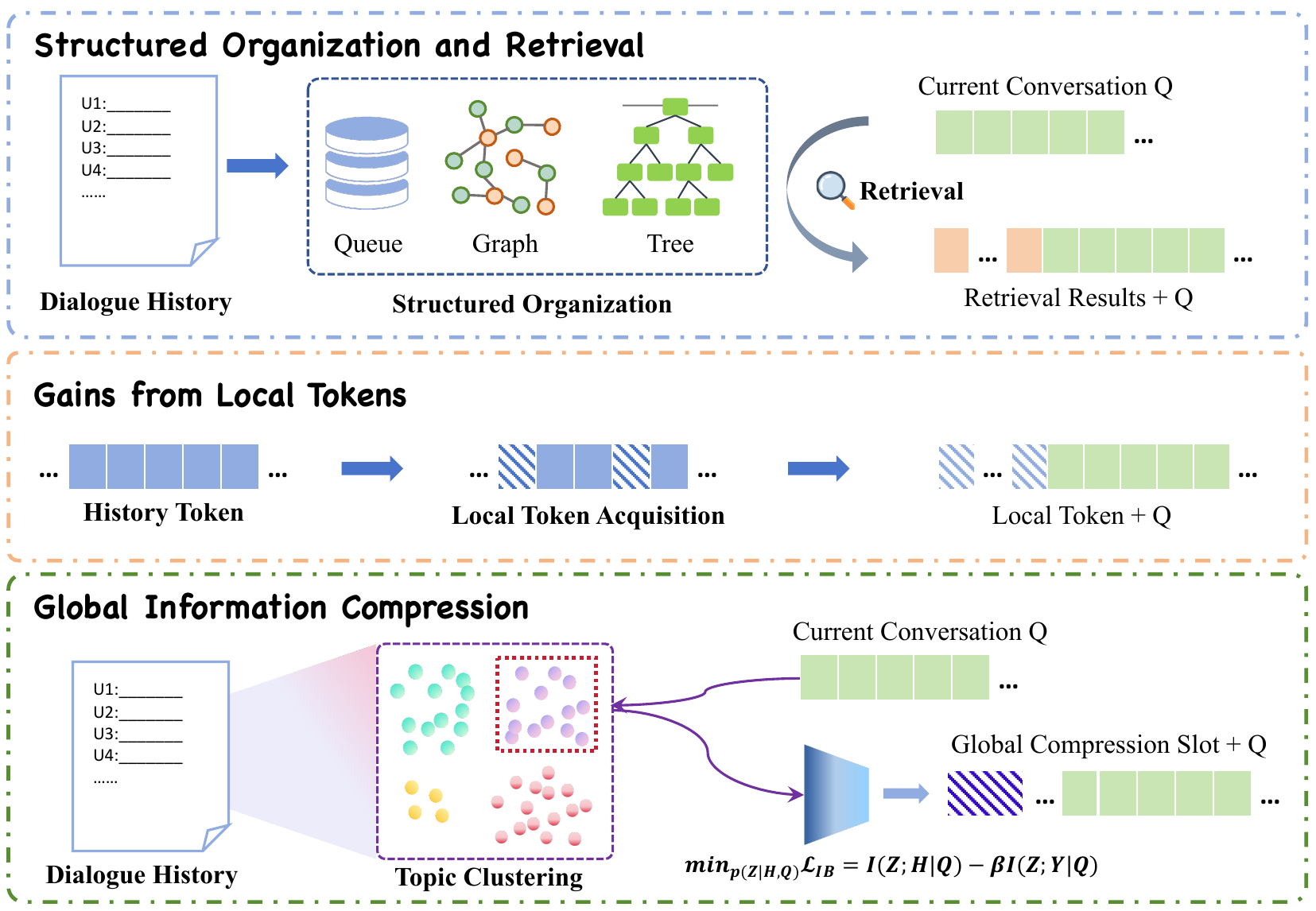}}
    \caption{
      Rigid Retrieval \& Local Tokens vs. History Compression in CoLink.
    }
    \label{fig:intro-differ-method}
\end{figure}

Current strategies to mitigate history compression fall into two categories, as illustrated in Fig.~\ref{fig:intro-differ-method}. (1) Structured Organization and Retrieval methods organize segmented history into memory banks \cite{zhong2024memorybank}, trees \cite{rezazadeh2024isolated,sarthi2024raptor}, or graphs \cite{jimenez2024hipporag,edge2024local,gutierrez2025rag} for retrieval. Despite enabling long-term storage, these methods often suffer from information loss during construction and retrieval noise during generation. (2) Local Token Modeling methods approximate full-history attention by retaining a sparse subset of tokens, such as attention sinks or recent windows \cite{xiao2023efficient,li2024streamingdialogue,chen2024core}. Although computationally efficient, they tend to be myopic, failing to capture global semantic dependencies and the latent thematic coherence across turns.

In contrast to these existing approaches, which focus on structural organization or local context approximation, insights from human cognition suggest an alternative paradigm for semantic organization that emphasizes abstraction, dynamic memory activation, and global coherence \cite{huitt2003information,atkinson1968human}. Rather than storing verbatim history, the human brain distills interactions into abstract representations and dynamically activates relevant long-term memories to ground new inputs through the interplay of the prefrontal cortex and hippocampus. This mechanism ensures global semantic associativity and enables efficient sense-making \cite{cai2022memory}.

Drawing on these insights, we propose CoLink, a novel framework that seamlessly integrates cognitive memory modeling with global information compression. CoLink operates on two key levels: (1) Adaptive associativity and activation. To simulate short-long memory associativity, CoLink organizes history into topic graphs that enable cross-semantic linking with external knowledge. Upon receiving a new dialogue turn, it identifies the relevant historical clusters and dynamically activates context-specific knowledge via an adaptive control mechanism, exploiting both semantic associations and structural linkage importance to retrieve highly pertinent facts. (2) Sense-making via compression. CoLink adopts the Information Bottleneck (IB) principle \cite{zhu2024information} to compactly represent the expansive history and activated knowledge. Guided by the objective of minimizing the mutual information (MI) between the compressed representations and the raw inputs while maximizing the MI with the target response, we derive a tri-part loss comprising compression, prediction, and reconstruction to effectively balance globality and conciseness. To guarantee extreme compression efficiency, CoLink competitively distills salient information into a limited set of latent slots. This mechanism directly abstracts compact memory and knowledge slots from the full contexts, yielding highly condensed representations optimally tailored for LLM generation.

To evaluate CoLink in complex domains requiring multi-expert reasoning, where existing benchmarks often fall short, we design a knowledge-enhanced, self-reflective multi-agent debate framework and construct \textbf{MMD}, a \textbf{m}ulti-turn \textbf{m}edical \textbf{d}ialogue dataset with $2370$ samples. Extensive experiments demonstrate that CoLink significantly enhances information retention and cross-turn consistency, achieving up to +2.59\% gains over baselines even under a 32$\times$ compression constraint. Furthermore, additional evaluations on both open-domain dialogue and specialized disease diagnosis datasets firmly validate the strong domain universality and zero-shot generalizability of our framework.
Our paper presents three main innovations:
\begin{itemize}[itemsep=2pt,topsep=2pt,parsep=2pt,leftmargin=*]
\item To the best of our knowledge, we are the first to formulate context abstraction in multi-turn dialogues through the lens of the Information Bottleneck principle. By integrating this objective with a competitive latent slot mechanism, our framework achieves efficient and effective compression of global content.
\item We propose a cross-level linking mechanism that bridges dialogue history with external knowledge bases. With both semantic associations and graph-structural reasoning, our method dynamically activates knowledge subgraphs that are strictly relevant to the current conversational context.
\item We conduct extensive evaluations across diverse datasets using a multifaceted validation approach, including main comparisons, component/loss ablations, and in-depth effectiveness analyses. The empirical results demonstrate that CoLink exhibits clear advantages in compression efficiency, information retention, and generation accuracy.

\end{itemize}

\section{Related Work}\label{sec:related}

\subsection{History Organization and Utilization}
Hierarchical structures and retrieval mechanisms mitigate redundancy by selectively retaining crucial history and enabling localization of relevant context. MemoryBank \cite{zhong2024memorybank} partitions history into chunks and retrieves relevant blocks via similarity search. RAPTOR \cite{sarthi2024raptor} and MemTree \cite{rezazadeh2024isolated} organize dialogue into hierarchical trees through recursive clustering or semantic aggregation, enabling retrieval at multiple granularities.  Graph-based approaches, such as GraphRAG \cite{edge2024local}, HippoRAG \cite{jimenez2024hipporag}, ArchRAG \cite{wang2025archrag}, KG2RAG \cite{zhu2025knowledge}, and HippoRAG2 \cite{gutierrez2025rag}, extract entity–relation graphs and leverage graph traversal to identify relevant substructures. While these methods improve scalability and retrieval efficiency, they primarily rely on sparse entity or structural cues, often discarding fine-grained contextual information. Moreover, they lack explicit modeling of inter-turn semantic dependencies, which limits holistic integration and global compression of the dialogue history.

\subsection{Exploiting Potential Gains from Local Tokens}
These Local Token Modeling methods are fundamentally predicated on the phenomenon of attention sparsity. Empirical observations indicate that LLMs consistently allocate disproportionate attention weights to a minimal subset of tokens, particularly the initial and the last few tokens. This phenomenon is known as the "Attention Sink" \cite{kang2025see,zhang2024tell}. Leveraging this insight, methods such as StreamLLM \cite{xiao2023efficient}, StreamingDialogue \cite{li2024streamingdialogue}, and LM-Infinite \cite{liu2024lost} persistently cache initial tokens, turn delimiters, or recent tokens to approximate full-history attention. Similarly, H2O \cite{zhang2023h2o} dynamically retains only a fraction of tokens with the highest accumulated attention scores. Chen et al. \cite{chen2024core} introduce core context-aware attention, which partitions long contexts into groups and preserves only a single representative token per group.

However, recent studies \cite{sun2024massive,gan2026unleashing} reveal that these sinks predominantly serve as structural biases to maintain attention stability rather than as repositories of core semantics. By confining the receptive field to local windows or sparse anchors, these methods inevitably fragment the historical continuity. Consequently, they inherently lack global perception, leading to degraded performance in tasks that necessitate deep, long-horizon understanding.

\subsection{Cognitive Memory Mechanisms}
Cognitive neuroscience has extensively investigated the mechanisms of memory retention, retrieval, and knowledge representation. Advanced cognitive abilities rely not merely on the synergy among various memory systems, but fundamentally on the high-level abstraction of information \cite{atkinson1968human, liang2025ai}. Regarding memory architecture, Squire’s taxonomy classically distinguishes declarative memory (comprising semantic facts and episodic events) from non-declarative memory (e.g., procedural skills) \cite{squire2004memory}. Furthermore, research posits that \textit{semantic networks} serve as the critical interface between sensory inputs and consolidated long-term knowledge \cite{popham2021visual}. During dialogue processing, cognitive theories suggest that novel linguistic stimuli trigger a complex complementary interaction \cite{atkinson1968human}: the hippocampus acts as a rapid episodic index, which subsequently activates long-term semantic knowledge in the neocortex via associative retrieval. These activated memories are then integrated within short-term working memory to facilitate coherent language generation.

Motivated by these insights, recent studies have explored brain-inspired architectures for dialogue systems. MGMR simulates the memory-stage theory utilizing memory graphs and rehearsal mechanisms \cite{cai2022memory}, while HippoRAG models dialogue history through graph structures to enhance retrieval \cite{jimenez2024hipporag}. Nevertheless, these approaches predominantly focus on memory storage structures and flow pathways, largely overlooking a fundamental characteristic of semantic memory: its high degree of abstraction. Consequently, they lack principled mechanisms to distill abstract semantic representations from redundant signals to prevent information overload.

\begin{figure}[]
    \centerline{\includegraphics[width=\linewidth]{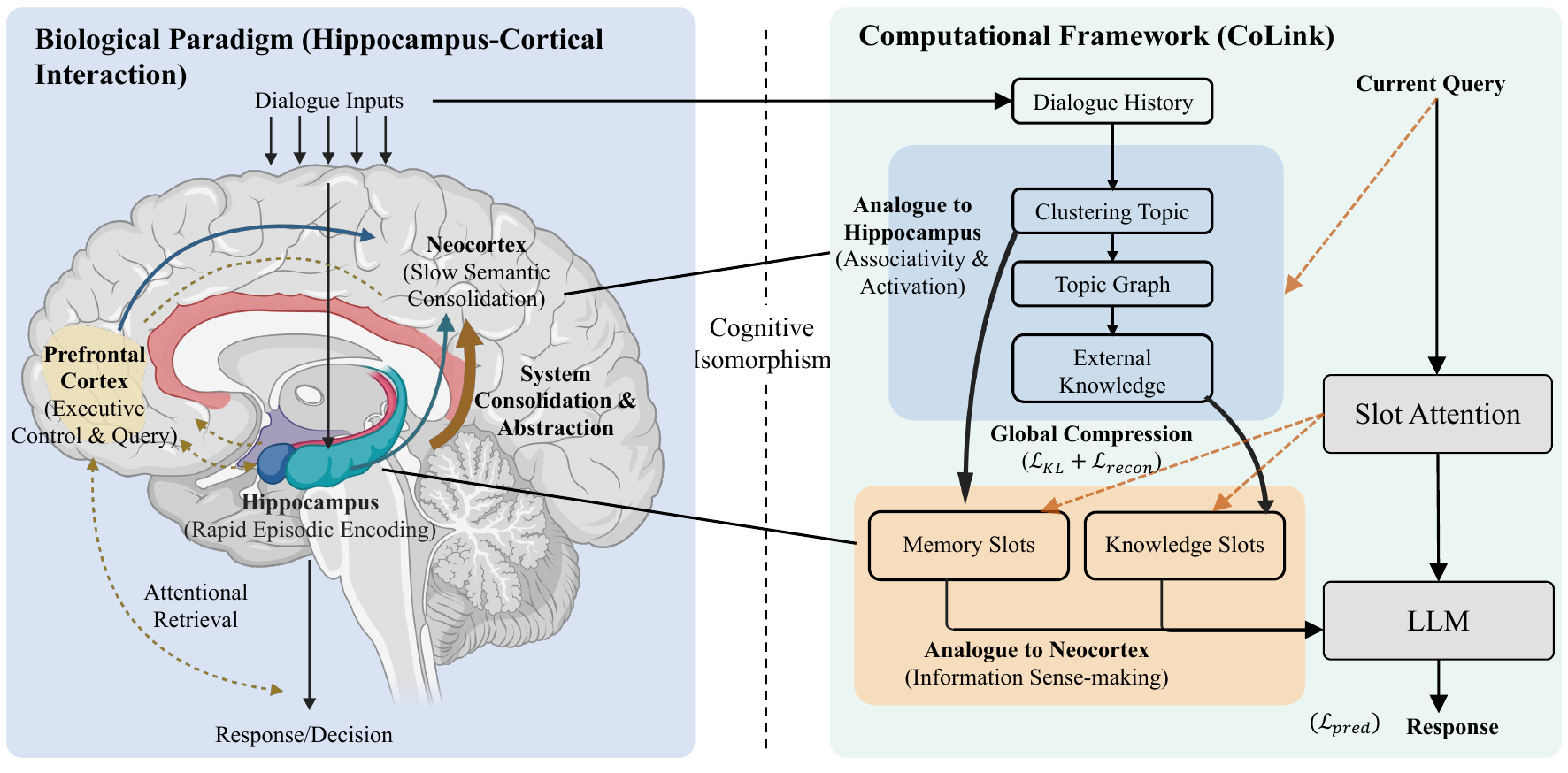}}
    \caption{
      Cognitive isomorphism between human memory mechanisms and the CoLink framework. The diagram illustrates two key biological parallels: (1) Associative Retrieval: Analogous to the hippocampus retrieving episodic memories to activate long-term cortical knowledge, CoLink constructs topic graphs to link historical contexts with external knowledge, dynamically activating relevant facts via the semantic relevance and structural linkage importance. (2) Abstract Encoding: Just as the human brain stores highly abstract semantic schemas rather than verbatim textual records, CoLink employs an global competitive compression strategy. This rigorously distills raw history and knowledge streams into compact Memory/Knowledge slots, mirroring the brain's efficiency in retaining core semantics while filtering redundancy.
    }
    \label{fig:cognitive-colink-framework}
\end{figure}

In this work, we bridge cognitive science and information theory to model multi-turn dialogues. We conceptualize the external knowledge base as long-term semantic memory. To simulate neural retrieval interactions, we construct a historical topic graph that functions as a hippocampal-like "memory index" to precisely activate context-relevant external knowledge. Crucially, we introduce IB-driven objectives to emulate the brain's inherent semantic abstraction process. This objective rigorously distills core information from complex, intertwined history and knowledge streams, ensuring that the response generation relies exclusively on the most critical semantic features. The conceptual diagram of CoLink is visually summarized in Fig.~\ref{fig:cognitive-colink-framework}.

\section{Theoretical Framework and Objective Derivation}\label{sec:theory}
In this section, we establish the foundational objectives and theoretical formulations of our approach. We first formalize the multi-turn dialogue task for LLMs and introduce the preliminaries of the Information Bottleneck (IB) principle \cite{tishby2000information}. Building upon this, we extend the IB framework to multi-turn dialogues and rigorously derive the theoretical objectives for the global compression of dialogue history and associated knowledge.

\subsection{Problem Definition}
For the multi-turn dialogue task, the objective of LLMs is to generate a response given the historical dialogue $H = [u_1, \dots, u_{t-1}]$ and the current query $Q = u_t$. Through autoregressive prediction, the model maximizes the joint distribution of the target sequence $Y = (y_1, y_2, \dots, y_m)$. Furthermore, considering that domain-specific conversations (e.g., healthcare) often require auxiliary background knowledge $K$, the objective can be formulated as:
\begin{equation}
    \mathcal{L}_{gen}=max \displaystyle\sum_{i=1}^m \log P(y_{i}|H,K,u_t,y_{<{i}}) \label{eq1}
\end{equation}
where $y_{<i} = (y_1, \dots, y_{i-1})$ denotes the previously generated tokens. While LLMs rely on attention mechanisms to model long-range dependencies, it remains challenging to assign precise attention weights to the most relevant historical and knowledge contexts as the dialogue progresses, amid extensive conversational noise.

\subsection{Information Bottleneck Preliminary}
The Information Bottleneck principle provides a rigorous framework for learning an intermediate representation $Z$ of input signals $H$ that preserves maximal information relevant to the target variable $Y$. The fundamental objective is shown in:
\begin{equation}
    min \mathcal{L}_{IB}=I(Z;H)-\beta I(Z;Y) \label{eq2}
\end{equation}
where $I(\cdot)$ denotes mutual information and $\beta$ is a Lagrange multiplier. For variables $h \in H$ and $z \in Z$, the mutual information term can be expressed via Kullback-Leibler (KL) divergence:
\begin{equation}
\resizebox{0.8\linewidth}{!}{$
    \begin{aligned} I(Z;H) &= D_{KL}[p(z,h) \| p(z) \otimes p(h)] \\ &= \int_z \int_h p(z,h) \log \frac{p(z,h)}{p(z)p(h)} \, dh \, dz \end{aligned} \label{eq3}
$}
\end{equation}
Effective modeling of long-horizon dialogue requires compressing extensive context to extract salient semantics while balancing the suppression of redundancy with the retention of long-term dependencies \cite{chen2024core}. The IB framework perfectly aligns with our requirement for global information compression during multi-turn interactions. Therefore, we adopt it as our theoretical foundation. We define the two components of the IB objective as compression (minimizing $I(Z;H)$) and prediction (maximizing $I(Z;Y)$), which conceptually strive to eliminate input redundancy while retaining the core semantics necessary for generation.

\subsection{Global Context Sense-Making Strategy}
Since historical text and external knowledge are inherently distinct, we design dual compressors—memory and knowledge—based on the same theoretical objective to ensure LLMs can effectively disentangle and utilize these abstract representations. In the following subsections, we detail the derivation of the global compression objectives for both the historical sequences and the related knowledge.

\subsubsection{Derivation of Memory Compression Objective}\label{sec:theory-3-3-1}
The compression of the history $H$ must be conditioned on the current query $Q$. For instance, in medical dialogues spanning diverse topics such as symptoms, lifestyle, and examinations, a query about a medication plan requires selectively integrating relevant historical context. Therefore, we formulate a conditional objective for history compression:
\begin{equation}
    \min_{P(Z|H,Q)} \mathcal{L}_{mem} = I(Z;H|Q) - \beta I(Z;Y|Q) \label{eq4}
\end{equation}

The first term enforces compactness by penalizing redundant information, while the second term ensures the preservation of salient information required for generation. We expand the \textbf{Compactness} term as:
\begin{equation}
\resizebox{0.9\linewidth}{!}{$
    \begin{aligned} I(Z;H|Q) &= \mathbb{E}_q \left[ \int_h \int_z p(z,h|q) \log \frac{p(z,h|q)}{p(z|q)p(h|q)} \, dz \, dh \right] \\ &= \mathbb{E}_q \left[ \int_h \int_z p(h|q)p(z|h,q) \log \frac{p(z|h,q)}{p(z|q)} \, dz \, dh \right] \\ &= \mathbb{E}_{p(h,q)} \left[ \int_z p(z|h,q) \log \frac{p(z|h,q)}{p(z|q)} \, dz \right] \\ &= \mathbb{E}_{p(h,q)} \left[ \mathbb{D}_{KL}[p(z|h,q) \| p(z|q)] \right] \end{aligned} \label{eq5}
$}
\end{equation}
Minimizing $I(Z;H|Q)$ compels the model to compress the history $H$ as aggressively as possible relative to $Q$, ensuring $Z$ should not depend on $H$.

The \textbf{Preservation} term is derived as:
\begin{equation}
\resizebox{0.9\linewidth}{!}{$
    \begin{aligned} I(Z;Y|Q) &= \mathbb{E}_q \left[ \int_y \int_z p(z,y|q) \log \frac{p(z,y|q)}{p(z|q)p(y|q)} \, dz \, dy \right] \\ &= \mathbb{E}_q \left[ \int_y \int_z p(z|q)p(y|z,q) \log \frac{p(y|z,q)}{p(y|q)} \, dz \, dy \right] \\ &= \mathbb{E}_{p(y,z,q)} \left[ \log \frac{p(y|z,q)}{p(y|q)} \right] \\ &= \mathbb{E}_{p(y,z,q)} \left[ \log p(y|z,q) \right] - \mathbb{E}_{p(y,q)} \left[ \log p(y|q) \right] \end{aligned} \label{eq6}
$}
\end{equation}

Since the second term in \eqref{eq6} is independent of the compression $Z$, $maxI(Z;Y|Q)$ is equivalent to $max\mathbb{E}_{p(y,z,q)}[\log p(y|z,q)]$. This ensures that $Z$ alone suffices to generate accurate responses, thereby guaranteeing the retention of the most task-relevant semantics.

We define the global history compression encoder as $q_{\phi}(z|h,q)$ and the LLM generator as $p_{\theta}(y|z,q)$. Since the marginal $p(z|q)$ is intractable, we optimize the variational upper bound using a parametric prior $r_\varphi(z|q)$. The final training objective for memory compression becomes:
\begin{equation}
    \begin{aligned}
\mathcal{L}_{mem} = & \ \mathbb{E}_{p(h,q)}\Big[ \mathbb{D}_{KL}[q_{\phi}(z|h,q) \| r_\varphi(z|q)] \Big] \\
& + \beta\mathbb{E}_{p(y,h,q)}\mathbb{E}_{z \sim q_{\phi}(\cdot|h,q)}\Big[ -\log p_{\theta}(y|z,q) \Big]
\end{aligned} \label{eq7}
\end{equation}

\subsubsection{Derivation of Knowledge Compression Objective}\label{sec:theory-3-3-2}
During dialogue, historical information extraction is independent of external knowledge and conditioned solely on the current query. In contrast, global modeling of relevant knowledge needs to additionally consider the dialogue state encoded in the compressed history. Let $K$ denote the set of retrieved knowledge paths activated by $Q$. The goal of the knowledge compressor is to learn a representation $L$ that captures the minimal sufficient statistics of $K$ required to generate $Y$, conditioned on $Q$ and $Z$. Since the derivation closely follows Section~\ref{sec:theory-3-3-1}, we omit the intermediate steps for brevity. The fundamental objective function is formulated as:
\begin{equation}
    \min_{P(L|K,Q,Z)} \mathcal{L}_{know} = I(L;K|Q) - \beta I(L;Y|Q, Z) \label{eq8}
\end{equation}

The \textbf{Compactness} term minimizes the mutual information between the compressed knowledge $L$ and the raw knowledge $K$, filtering out irrelevant noise. Using a variational approximation with a conditional prior $r_\varphi(l|q)$ and a knowledge compression encoder $q_{\psi}(l|k,q)$, we obtain the upper bound:
\begin{equation}
    I(L;K|Q) \leq \mathbb{E}_{p(k,q)} \left[ \mathbb{D}_{KL}[q_{\psi}(l|k,q) \| r_\varphi(l|q)] \right] \label{eq9}
\end{equation}

The \textbf{Preservation} term for knowledge is derived as:
\begin{equation}
\resizebox{\linewidth}{!}{$
    \begin{aligned} I(L;Y|Q, Z) &= H(Y|Q, Z) - H(Y|L, Q, Z) \\ &= H(Y|Q, Z) + \mathbb{E}_{p(k,y,q,z)} \left[ \mathbb{E}_{l \sim q_{\psi}(\cdot|k,q)} [\log p_{\theta}(y|l, z, q)] \right] \end{aligned} \label{eq10}
$}
\end{equation}
Since $H(Y|Q, Z)$ is constant with respect to the compressor parameters $\psi$, maximizing this mutual information is equivalent to maximizing the expected log-likelihood of the generator $p_{\theta}$. Thus, the final knowledge compression objective is formulated as:
\begin{equation}
    \begin{aligned}
\mathcal{L}_{know} = & \ \mathbb{E}_{p(k,q)} \left[ \mathbb{D}_{KL}[q_{\psi}(l|k,q) \| r_\varphi(l|q)] \right] \\
& +\beta \mathbb{E}_{p(k,y,q,z)} \left[ \mathbb{E}_{l \sim q_{\psi}(\cdot|k,q)} [\log p_{\theta}(y|l, z, q)] \right]
\end{aligned} \label{eq11}
\end{equation}

Derived from \eqref{eq7} and \eqref{eq11}, we obtain the compressed representations $Z$ and $L$ ($Z \ll H$ and $L \ll K$). Consequently, the original generative objective in \eqref{eq1} is reformulated as:
\begin{equation}
    \mathcal{L}_{gen}=max \displaystyle\sum_{i=1}^m \log P(y_{i}|Z,L,u_t,y_{<{i}}) \label{eq12}
\end{equation}

In Section~\ref{sec:method}, we further describe the adaptive activation of external knowledge, alongside the concrete instantiation and training objectives of the compressors.

\section{Method}\label{sec:method}
\begin{figure*}[]
    \centerline{\includegraphics[width=0.95\linewidth]{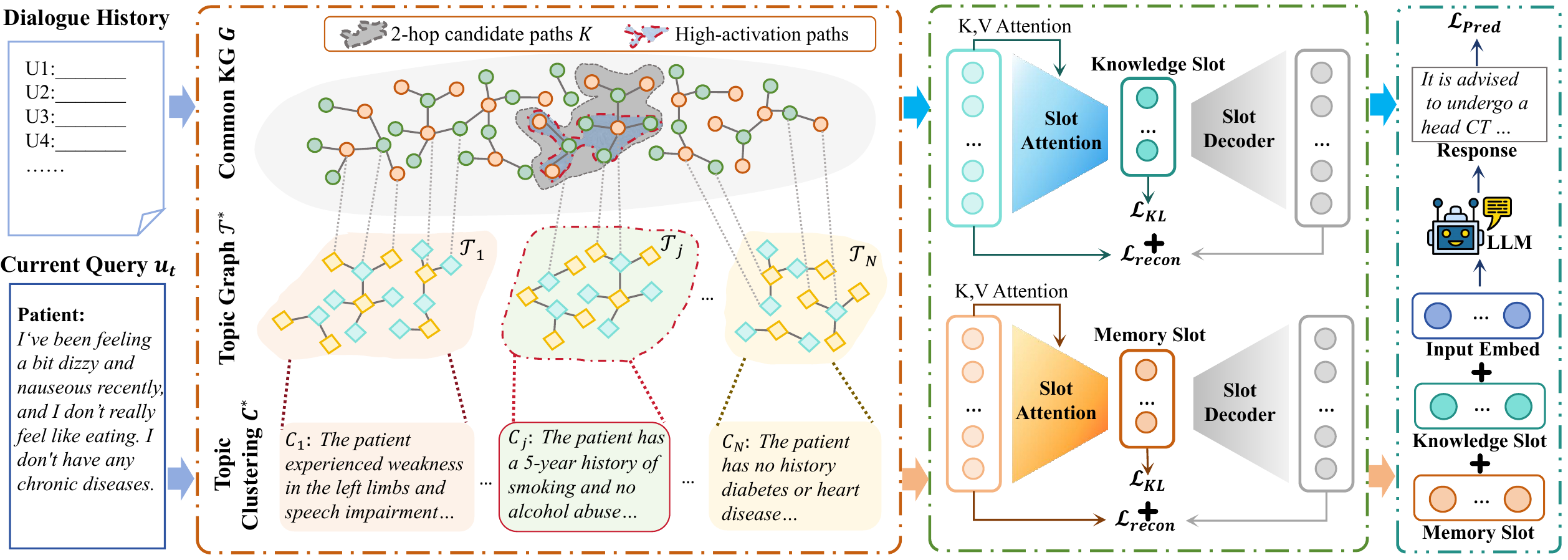}}
    \caption{
      The framework of CoLink.
    }
    \label{fig:colink-framework}
\end{figure*}
In this section, we detail the architectural pipeline and implementation of CoLink. Inspired by cognitive sense-making and associativity, CoLink integrates memory modeling with information-theoretic compression. As described in Section~\ref{sec:me-association}, it constructs historical topic graphs to establish cross-semantic links with external knowledge and dynamically activates context-relevant facts by exploiting the semantic and structural associations of structured knowledge. To guarantee efficient global compression and the retention of core semantics from long contexts, CoLink competitively abstracts raw inputs into a small set of latent slots, namely \textit{memory slots} and \textit{knowledge slots} (Section~\ref{sec:me-slot}). In Section~\ref{sec:me-obj}, we design a tri-part training loss, comprising KL divergence, reconstruction, and prediction, to enforce global information compression and semantic alignment under the theoretical formulations derived in Section~\ref{sec:theory}. An overview of the framework is illustrated in Fig.~\ref{fig:colink-framework}.

\subsection{Association and Activation}\label{sec:me-association}
\subsubsection{Cross-Level Semantic Linking}\label{sec:me-cross}
To preserve dialogue coherence and global consistency while mitigating interference across distinct topics, CoLink clusters the dialogue history $H=[u_1,\dots,u_{t-1}]$ into thematic clusters $C^*=[C_1,\dots,C_N]$ using BERTopic \cite{grootendorst2022bertopic}. For each cluster, we apply OpenIE to extract relational triples and construct a set of topic graphs $\mathcal{T}^*=\{\mathcal{T}_1,\dots,\mathcal{T}_N\}$. Each graph $\mathcal{T}_i=(E^i,R^i)$ encodes the entities $E^i$ and relations $R^i$ associated with the $i$-th theme. For knowledge augmentation, these local graphs are then grounded to the external Knowledge Graph (KG) $G=(E,R)$ via entity alignment to construct a multi-layered semantic bridge. The alignment is achieved by maximizing the cosine similarity of the entities from the topic graph and $G$:
\begin{equation}
    \max Sim(e^{\mathcal{T}},e^G)=\frac {\mathbf{e}^{\mathcal{T}} \cdot \mathbf{e}^G}{||\mathbf{e}^{\mathcal{T}}||\cdot||\mathbf{e}^G||} \label{eq13}
\end{equation}
where $e^{\mathcal{T}}$ and $e^G$ are the entities from the topic graph $\mathcal{T}$ and the knowledge graph $G$, respectively, while $\mathbf{e}^{\mathcal{T}}$ and $\mathbf{e}^G$ represent their corresponding semantic embeddings.

As the dialogue progresses, both the thematic clusters and their corresponding topic graphs expand dynamically. This structure functions as a "semantic episodic memory index," providing the foundational support for subsequent knowledge activation and memory compression.

\subsubsection{Adaptive Control of Knowledge Activation}\label{sec:me-activation}
When a new user utterance $u_t$ arrives and is assigned to a historical topic cluster $C_j$, CoLink employs adaptive control of thought to activate context-relevant knowledge paths in the external KG. Given a set of $2$-hop candidate paths $K=[k_1,\dots,k_N]$, where the paths are constructed from entities in the external KG $G$ aligned with the historical topic graph $\mathcal{T}_j$, the activation score of path $k_k$ is determined by its semantic relevance to the dialogue entities and its structural linkage importance, as defined in \eqref{eq14}:
\begin{equation}
    A_k=B_k+\sum_{e \in E^{\mathcal{T}_j} \cap E^{k_k}} w_jS_{jk} \label{eq14}
\end{equation}
where $B_k$ denotes the baseline activation derived from the semantic similarity between $k_k$ and $u_t$. $E^{\mathcal{T}_j}=[e^{\mathcal{T}_j}_1, \dots, e^{\mathcal{T}_j}_{|E^{\mathcal{T}_j}|}]$ and $E^{k_k}=[e^{k_k}_1, e^{k_k}_2, e^{k_k}_3]$ denote the entity sets of the topic graph $\mathcal{T}_j$ and the path $k_k$, respectively. The term $w_j$ is a time-decay factor that downweights entities associated with temporally distant utterances. Assuming $E^{G}=[e_1,\dots,e_{|E^{\mathcal{T}_j}|}]$ represents the set of external knowledge entities aligned with $E^{\mathcal{T}_j}$, the variable $S_{jk}$ quantifies the associative strength between the aligned entity $e_j \in E^{G} \cap E^{k_k}$ and the path $k_k$. Details of the computation are provided below.

For each node $e_j \in E^{G} \cap E^{k_k}$, we isolate the connected subgraph $G^{'}=(E^{'},R^{'})$ anchored at node $e_j$. We then initialize the graph-structural relevance $\mathbf{S}_j$ between all subgraph nodes and $e_j$ using a Gaussian kernel, and iteratively update it via Random Walk with Restart (RWR) \cite{liu2025efficient} over the upper-layer KG until convergence. The initialization and the update rule are defined as: 
\begin{equation}
    \begin{aligned}
S^{(0)}_{ji} = exp(- \frac{dis(j, i)^2}{2h^2}), where e_i \in E^{'} \\ \mathbf{S}_j^{'} = c \cdot \tilde{\mathbf{A}} \mathbf{S}_j + (1-c) \cdot \mathbf{O}_j \label{eq15} 
    \end{aligned}
\end{equation}
where $h$ is the bandwidth and $\tilde{\mathbf{A}} \in \mathbb{R}^{|E^{'}| \times |E^{'}|}$ is the transition matrix of the knowledge subgraph. In \eqref{eq15}, $\mathbf{O}_j$ is the one-hot restart vector centered at the entity $e_j$, and $c$ denotes the restart probability. The resulting vector $\mathbf{S}_j^{'}$ captures the global graph relevance of entity $e_j$ to all entities in the subgraph $G^{'}$, from which the specific associativity $S_{jk}$ for the path $k_k$ is extracted:
\begin{equation}
    S_{jk}=\max_{e^{k_k} \in E^{k_k}} \mathbf{S}^{'}_j(e^{k_k}) \label{eq16}
\end{equation}

Ultimately, all entities aligned with $\mathcal{T}_j$ jointly determine the activation level $A_k$.

Structurally, this framework mirrors the hippocampus-cortex interplay: the topic graph functions as a dynamic contextual index, while the external KG serves as consolidated long-term storage \cite{atkinson1968human, jimenez2024hipporag, cai2022memory}. Their interaction through context-adaptive activation facilitates efficient semantic integration and retrieval.

\subsection{Competitive Extraction of Salient Information} \label{sec:me-slot}
To instantiate the theoretical variables $Z$ and $L$, we propose abstracting the raw, variable-length representations of global history $H$ and knowledge $K$ into a fixed, highly condensed set of latent bottleneck slots. This architectural design natively guarantees the strict compactness bounds ($Z \ll H$ and $L \ll K$) while preserving seamless compatibility with LLM generators. To operationalize this competitive abstraction, we adapt the Slot Attention \cite{locatello2020object} as our computational backbone, which iteratively distills complex long-range dependencies into these fixed-capacity slots. Since the memory and knowledge compressors use the same training framework, we describe the historical compressor as an example. Specifically, we initialize $M$ learnable memory slots $\mathbf{S}^0 = [s^0_1, \dots, s^0_M] \in \mathbb{R}^{M \times d}$ as the concrete instantiation of $Z$ to distill the salient information.

For history compression, let $C_h$ denote the relevant historical cluster. We encode $C_h$ into a sequence representation $\mathbf{X}=[x_1, \dots, x_L] \in \mathbb{R}^{L \times d}$
. First, the inputs and slots are projected into key, value, and query spaces via linear transformations and Layer Normalization (LN) \cite{ba2016layer}:
\begin{equation}
\resizebox{0.9\linewidth}{!}{$
    \mathbf{K}=\mathbf{W}_k \text{LN}(\mathbf{X}), \quad \mathbf{V}=\mathbf{W}_v \text{LN}(\mathbf{X}), \quad \mathbf{Q}=\mathbf{W}_q \text{LN}(\mathbf{S}^t) \label{eq17}
$}
\end{equation}

where $\mathbf{W}_k, \mathbf{W}_v, \mathbf{W}_q$ are learnable projections. The slots then perform a selective aggregation of input features through softmax attention and obtain the updates $\widetilde{\mathbf{U}}^t$:

\begin{equation}
    \begin{aligned}
    \mathbf{A}^t_{i,j} &= \frac{\exp(\mathbf{Q}_i \cdot \mathbf{K}j / \sqrt{d})}{\sum_{k=1}^M \exp(\mathbf{Q}_k \cdot \mathbf{K}_j / \sqrt{d})} \\
    \widetilde{\mathbf{U}}^t &= \text{Norm}(\mathbf{A}^t) \cdot \mathbf{V}, \quad \text{Norm}(\mathbf{A}^t)_{i,j} = \frac{\mathbf{A}^t_{i,j}}{\sum_{l=1}^L \mathbf{A}^t_{i,l}}  \end{aligned}\label{eq18}
\end{equation}
The aggregated updates are fused into the slots using a Gated Recurrent Unit (GRU) \cite{dey2017gate}, as described in \eqref{eq18}. After $T$ iterations, the resulting slots $\mathbf{S}^T$ constitute the compressed latent representation $Z$.

\begin{equation}
    \begin{aligned} \widetilde{\mathbf{S}}^t &= \text{GRU}(\widetilde{\mathbf{U}}^t, \mathbf{S}^{t-1}) \\ \mathbf{S}^{t} &= \widetilde{\mathbf{S}}^t + \text{MLP}(\text{LN}(\widetilde{\mathbf{S}}^t)) \end{aligned} \label{eq19}
\end{equation}
Since each slot interacts with all tokens through attention, the module captures globally salient information while achieving an effective compression ratio of $|H|/M$. By prepending the memory slots and knowledge slots as prefix tokens for LLMs, CoLink effectively enhances the models' perception of critical context.

\subsection{Training Objectives and Optimization}\label{sec:me-obj}
To effectively optimize the compressors, we operationalize the objective derived in Section~\ref{sec:theory} via a composite loss function comprising three terms, as defined in \eqref{eq20}.

\textbf{KL Divergence Loss ($\mathcal{L}_{KL}$):} Constrains the posterior $q_{\phi}(z|h,q)$ to approximate the prior $r_\varphi(z|q)$, enforcing aggressive compression of the history relative to the query.

\textbf{Prediction Loss ($\mathcal{L}_{pred}$):} Maximizes the log-likelihood of the ground-truth response $y$, ensuring the compressed slots retain sufficient utility for generation.

\textbf{Reconstruction Loss ($\mathcal{L}_{rec}$):} Strictly minimizing mutual information can sometimes lead to representation collapse. To preserve interpretability and recoverability, we incorporate an auxiliary objective that reconstructs distilled context summaries $\hat{h}$ (via GPT-4 \cite{achiam2023gpt}).
\begin{equation}
    \begin{aligned}
\mathcal{L}_{mem} &= \underbrace{\alpha\mathbb{D}_{KL}[q_{\phi}(z|h,q)\|r_\varphi(z|q)]}_{\text{Compression}} \\
&\quad + \underbrace{\beta\mathbb{E}_{z}[-\log p_{\theta}(y|z,q)]}_{\text{Prediction}} \\
&\quad + \underbrace{\lambda \mathbb{E}_{z}[-\sum \log p(\hat{h}_t|\hat{h}_{<t},z)]}_{\text{Reconstruction}}
\end{aligned} \label{eq20}
\end{equation}
where $\alpha, \beta, \lambda$ are hyperparameters balancing each term. The objective for the knowledge compressor follows an analogous formulation as \eqref{eq21}.

\begin{equation}
    \begin{aligned} \mathcal{L}_{knowledge} &= \underbrace{\alpha \mathbb{E}_{p(k,q)} \left[ \mathbb{D}_{KL}[q_{\psi}(l|k,q) \| r_\varphi(l|q)] \right]}_{\text{Compression}} \\ &\quad + \underbrace{\beta \mathbb{E}_{p(y,k,q,z)} \mathbb{E}_{l \sim q_{\psi}} \left[ -\log p_{\theta}(y|l,z,q) \right]}_{\text{Prediction}} \\ &\quad + \underbrace{\lambda \mathbb{E}_{l \sim q_{\psi}} \left[ -\sum_{t} \log p(\hat{k}_t|\hat{k}_{<t}, l) \right]}_{\text{Reconstruction}} \end{aligned} \label{eq21}
\end{equation}
where $\hat{k}$ represents the target knowledge summary.

To ensure training stability and semantic consistency with the backbone LLM, we adopt a two-stage training strategy. In stage 1, we freeze the LLM parameters and optimize only the compression modules. In stage 2, we employ LoRA \cite{hu2022lora} to jointly fine-tune both the compressors and the LLM.

\section{Experiment}\label{sec:experiment}

\subsection{Datasets}\label{sec:ex-data}
\begin{figure*}[]
  \vskip 0.2in
  \begin{center}
    \centerline{\includegraphics[width=0.75\linewidth]{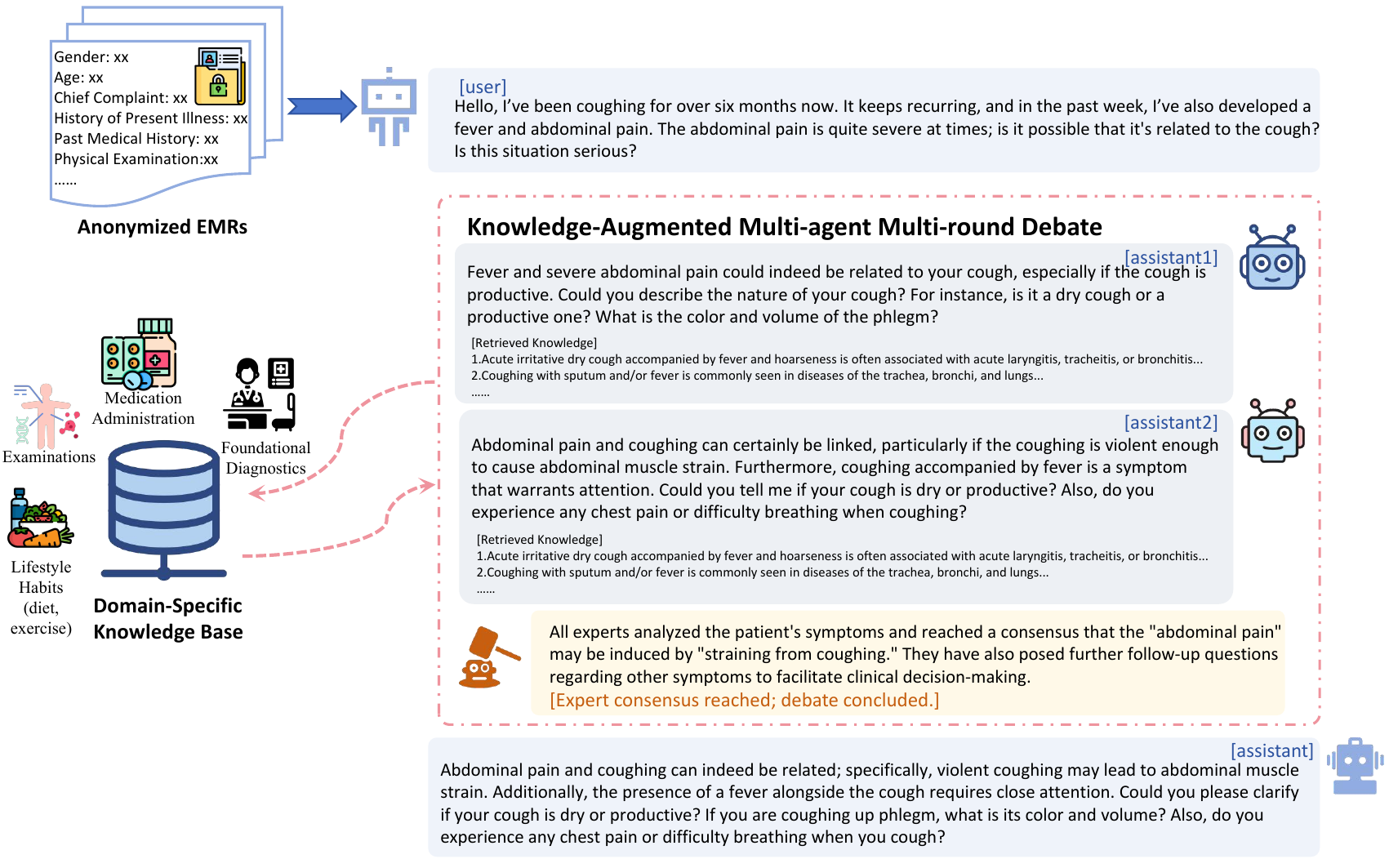}}
    \caption{
      Example of the data construction process.
    }
    \label{fig:data-con-exa}
  \end{center}
\end{figure*}

Existing medical dialogue datasets often suffer from knowledge leakage, role confusion, poor coherence, and factual hallucinations~\cite{sun2024parrot,kong2024platolm,wang2025learning}. To mitigate these issues and enhance granularity, we construct the MMD dataset using a novel knowledge-augmented multi-agent debate framework.
This framework simulates expert consultations involving three roles: User, Domain Experts ($N \ge 2$), and Judge.
\begin{itemize}
    \item User Agent: Initiates queries related to specific diseases.
    \item Expert Agent Group: Each expert performs reasoning based on a shared domain knowledge base, retrieves relevant facts, reflects on retrieved content, and filters out irrelevant knowledge.
    \item Judge Agent: Evaluates and aggregates expert opinions, and outputs a final diagnosis or recommendation only when a consensus is reached.
\end{itemize}
Dialogue scenarios follow real medical settings (e.g., diagnosis, examination recommendation, medication). Each conversation is grounded in an anonymized electronic health record and a corresponding knowledge base. For each user query, experts, augmented by a private, expert-only knowledge base, engage in iterative debate to correct errors. The final response is emitted only upon consensus verified by the Judge agent (see Fig.~\ref{fig:data-con-exa}).

Our primary dataset, MMD, encompasses $2370$ samples across $37$ diseases, with an average of $2.77$ debate rounds per response. The train/test split is $9:1$. Three general practitioners, each with over three years of clinical experience, evaluate knowledge relevance, debate logic, and response accuracy using a 5-point scale. Among $800$ samples, $94\%$ are rated as high quality, and inter-annotator agreement reaches a Cohen’s Kappa of $0.86$. 
To assess generalization, we further evaluate on the public benchmark CMtMedQA \cite{yang2024zhongjing}. Statistics for both MMD and CMtMedQA are summarized in Table~\ref{tab:dataset_statistics}.
\begin{table}[t]
    \centering
    \caption{Dataset statistics.}
    \label{tab:dataset_statistics}
    \resizebox{\linewidth}{!}{
        \begin{tabular}{lccccccc}
            \toprule
            \textbf{Dataset} & \textbf{Domain} & \textbf{Lang.} & \textbf{\# Samples} & \textbf{\# Diseases} & \textbf{Avg. Turns} & \textbf{Debate Turns} \\
            \midrule
            MMD & Medical & Chinese & 2,370 & 37 & 8.09 & 2.77 \\
            CMtMedQA & Medical & Chinese & 5,993 & 14 & 8.85  & -- \\
            \bottomrule
        \end{tabular}
    }
\end{table}

\subsection{Baselines}\label{sec:ex-baseline}
We perform extensive comparisons against standard context strategies and two primary paradigms.

\textbf{Standard Baselines:} We report performance using full history concatenation and summaries of all history.

\textbf{Structured Organization \& Retrieval:} We evaluate MemoryBank \cite{zhong2024memorybank} , MemTree \cite{rezazadeh2024isolated} and RAPTOR \cite{sarthi2024raptor}, and graph-based methods including GraphRAG \cite{edge2024local}, HippoRAG \cite{jimenez2024hipporag}, and HippoRAG2 \cite{gutierrez2025rag}. They all use OpenAI encoders for semantic-similarity-based retrieval. We also include NV-Embed \cite{lee2024nv} as a dense retrieval baseline. 

\textbf{Local Token Attention:} We compare against StreamLLM \cite{xiao2023efficient} and CCA \cite{chen2024core}.

All baselines are implemented following their original papers, including the information organization schemes and retrieval depths.
 \begin{table}[h]
\centering
    \caption{Hyperparameter settings for CoLink.}
    \label{tab:hyperparameters}
    \resizebox{0.95\linewidth}{!}{
        \begin{tabular}{ccc}
        \toprule
        \textbf{Hyperparameter} & \textbf{Description} & \textbf{Value} \\
        \midrule
        $E_1$ & Stage 1 Epochs & $10$ \\
        $E_2$ & Stage 2 Epochs & $10$ \\
        $\eta_1$ & Stage 1 Learning Rate & $5e-5$ \\
        $\eta_2$ & Stage 2 Learning Rate & $1e-5$ \\
        $w_d$ & Weight Decay & $0.01$ \\
        $B$ & Batch Size & $4$ \\
        $\mathbf{X}$ & Input Representation & Last Hidden State \\
        $d$ & Hidden Dimension & $1536$ \\
        $T$ & Slot Iterations & $3$ \\
        $\alpha$ & Compression Coeff. & $1$ \\
        $\beta$ & Prediction Coeff. & $1$ \\
        $\lambda$ & Reconstruction Coeff. & $1$ \\
        \bottomrule
\end{tabular}
}
\end{table}
\subsection{Experimental Details}\label{sec:ex-detail-evaluation}
We implement CoLink using PyTorch and optimize it via the Adam optimizer. All experiments are conducted on a single NVIDIA A800 (80GB) GPU. To ensure semantic alignment between the compressor and the generator, we initialize the input $\mathbf{X}$ using the final-layer hidden states of the backbone LLM. During inference, we set the temperature to $0$ and the maximum sequence length to $2048$. For external knowledge integration, since the MMD and CMtMedQA datasets lack natively aligned external knowledge bases, we employ WIMedKG \cite{li2025quality} and CMeKG \cite{byambasuren2019preliminary} as their respective supporting KGs. Detailed hyperparameters are listed in Table~\ref{tab:hyperparameters}. All reported results are averaged over three independent runs. Regarding evaluation, we employ BLEU, ROUGE, and BERTScore as automated metrics. Furthermore, to explicitly assess dialogue consistency and correctness, we sample $200$ instances for both LLM-as-a-Judge and human evaluation.

\begin{table*}[ht]
\centering
\caption{The comparative results. We highlight the best performance. B, R, and BS stand for BLEU, Rouge, and Bertscore, respectively.}
\label{tab:main_result}
\resizebox{\textwidth}{!}{%
\begin{tabular}{lcccccccccccccccc}
\toprule
\multicolumn{1}{c}{\multirow{2}{*}{\textbf{Method}}} & \multicolumn{8}{c}{\textbf{MMD} (\%)}                                                                                                                                              & \multicolumn{8}{c}{\textbf{CMtMedQA} (\%)}                                                                                                                                             \\ \cmidrule(lr){2-9} \cmidrule(lr){10-17}
                        & \textbf{B1} & \textbf{B2} & \textbf{B3}        & \textbf{B4} & \textbf{R1} & \textbf{R2} & \textbf{RL} & \textbf{BS} & \textbf{B1}              & \textbf{B2}              & \textbf{B3}              & \textbf{B4}              & \textbf{R1}              & \textbf{R2}              & \textbf{RL}              & \textbf{BS}       \\ \midrule
MemoryBank              & 19.41                  & 8.55                   & 4.47      & 2.32                   & 23.31                   & 5.24                   & 14.98                  & 68.83                          & 27.20           & 14.20           & 8.34            & 4.90            & 29.69           & 9.06            & 21.67           & 73.75           \\
NV-embed                & 17.77                  & 8.10                   & 4.22      & 2.31                   & 23.34                  & 5.60                   & 15.35                  & 68.22                         & 23.82           & 12.18           & 7.14            & 4.19            & 27.57           & 8.17            & 20.09           & 72.10           \\
StreamLLM & 0.38                   & 0.06                   & 0.03      & 0.02                   & 1.75                   & 0.02                   & 1.69                   & 45.97                         & 0.60            & 0.11            & 0.06            & 0.04            & 3.70            & 0.21            & 0.99            & 46.08           \\
GraphRAG                & 16.45                  & 6.64                   & 3.24      & 1.65                   & 21.57                  & 4.37                   & 14.15                  & 67.91                         & 24.10           & 12.02           & 6.90            & 3.94            & 27.25           & 7.64            & 19.45           & 72.41           \\
CCA                     & 8.62                   & 2.99                   & 1.46      & 0.78                   & 13.61                  & 2.07                   & 8.75                   & 64.14                         & 4.14            & 1.32            & 0.56            & 0.27            & 11.39           & 1.49            & 7.98            & 58.67           \\
MemTree                 & 12.42                  & 4.66                   & 2.29      & 1.26                   & 18.43                  & 3.28                   & 13.51                  & 64.36                         & 13.97           & 4.75            & 2.33            & 1.24            & 16.10           & 2.59            & 11.95           & 64.31           \\
RAPTOR                  & 14.91                  & 7.44                   & 4.18      & 2.50                   & 23.87                  & 6.77                   & 16.86                  & 68.74                         & 22.37           & 12.00           & 7.42            & 4.61            & 28.46           & 9.14            & 22.10           & 72.40           \\
HippoRAG                & 8.14                   & 3.13                   & 1.51      & 0.78                   & 18.42                  & 3.05                   & 13.14                  & 64.43                         & 16.14           & 7.12            & 3.64            & 1.97            & 24.04           & 5.45            & 17.94           & 68.65           \\
HippoRAG2               & 16.31                  & 7.89                   & 4.42      & 2.49                   & 19.59                  & 5.18                   & 13.65                  & 64.81                         & -- & -- & -- & -- & -- & -- & -- & -- \\ \midrule
Base                    & 17.62                  & 6.82                   & 3.34      & 1.79                   & 21.65                  & 3.80                   & 13.42                  & 67.89                         & 18.73           & 9.16            & 5.20            & 2.94            & 24.95           & 6.45            & 16.09           & 71.45           \\
\; + Full History      & \underline{22.47}                  & \underline{11.04}                  & \underline{6.32}      & \underline{3.67}                   & \underline{27.19}                  & \underline{6.95}                   & 16.53                  & \underline{70.26}                         & \underline{29.42}           & \underline{16.52}           & \underline{10.45}           & \underline{6.55}            & \underline{33.86}  & \underline{11.45}           & 23.10           & \textbf{75.90}  \\
\; + Summary History   & 13.97                  & 6.41                   & 3.59      & 2.16                   & 26.74                  & 6.75                   & \underline{17.20}                  & 69.16                         & 25.21           & 13.98           & 8.84            & 5.51            & 33.19           & 11.11           & \underline{24.31}  & 74.24           \\
CoLink(Ours)            & \textbf{27.52}        & \textbf{13.87}        & \textbf{7.66} & \textbf{4.10}                 & \textbf{31.63}        & \textbf{8.64}        & \textbf{19.26}        & \textbf{72.84}                & \textbf{31.62} & \textbf{18.02} & \textbf{11.74} & \textbf{7.62} & \textbf{34.28} & \textbf{12.31} & \textbf{25.50} & \underline{75.36}           \\ \bottomrule
\end{tabular}
}
\end{table*}
\subsection{Main Analysis}\label{sec:ex-main}
\subsubsection{Comparison Results}\label{sec:ex-com}
Table~\ref{tab:main_result} presents the comparative performance across all datasets. All experiments utilize Qwen2.5-1.5B \cite{bai2023qwen} as the backbone model. Based on these results, we summarize the key findings as follows:

\textbf{CoLink attains effective global information compression.} CoLink consistently outperforms both structured retrieval and local token attention baselines, achieving an average gain of $2.59\%$ even under a 32$\times$ compression constraint across two datasets. This indicates its superior ability to compress long-context information. Notably, CoLink matches or even exceeds the performance of Full History. This validates that, governed by the Information Bottleneck principle, CoLink effectively distills sufficient statistics from the global context, filtering out redundancy and noise that may distract the full-context baseline while retaining decision-critical semantics.

\textbf{Rigid structuring compromises contextual integrity.} Although chunk-based retrieval methods, such as MemoryBank and NV-Embed, offer moderate improvements over the base model, rigid graph and tree-based approaches (e.g., GraphRAG, MemTree) often degrade performance. Our analysis reveals that while structured extraction preserves discrete entities, it fails to capture the holistic semantic flow of the text. Furthermore, entity retention is highly sensitive to extraction quality and provides limited benefits for long-range dependencies (see Section~\ref{sec:ex-case}). Similarly, Summary History, despite reducing redundancy, suffers from inevitable information loss during condensation and obstructs performance. In contrast, CoLink resolves this dilemma via gradient-optimized slots, proving that soft, learnable compression is a more robust strategy for retaining holistic semantics than rigid structuring.

\begin{table*}[ht]
\centering
\caption{Comparison of online resource consumption and compression ratios.}
\label{tab:com-ratio}
\resizebox{0.95\textwidth}{!}{%
\begin{tabular}{lcccccccccccccccc}
\toprule
\multicolumn{1}{c}{\multirow{2}{*}{\textbf{Method}}} & \multicolumn{3}{c}{\textbf{MMD}}                                                                                                                                                                         & \multicolumn{3}{c}{\textbf{CMtMedQA}}                                                                                                                  \\ \cmidrule(lr){2-4} \cmidrule(lr){5-7}
                        & \multicolumn{1}{c}{\textbf{History/Knowledge  Processing Time} (ms)} & \multicolumn{1}{c}{\textbf{GPU Memory} (MB)} & \textbf{Compression Ratio} & \textbf{History/Knowledge  Processing Time} (ms) & \textbf{GPU Memory} (MB)                       & \textbf{Compression Ratio}       \\ \midrule
MemoryBank              & 2113.57                                        & /                                   & 0.6388            & 4521.04                    & /           & 0.4884            \\
NV-embed                & 218.41                                         & 27674.82                          & 0.8095            & 202.69                     & 27612.20            & 0.7732            \\
StreamLLM   & /                                                & 3050.95                           & 0.9665            & /                            & 3064.59             & 1.0000            \\
GraphRAG                & 41511.27                                       & /                                   & 2.8946            & 47679.49                   & /                     & 2.4866            \\
CCA                     & /                                                & 3679.53                           & 0.8903            & /                            & 3724.54             & 1.0000            \\
MemTree                 & 51725.50                                       & /                                   & 1.7233            & 3911.10                    & /                     & 1.5834            \\
RAPTOR                  & 763.50                                         & /                                   & 0.9900            & 823.20                     & /                     & 0.9900            \\
HippoRAG2               & 9149.56                                        & /                                   & 0.8328            & 10195.56                   & /                     & 0.9005            \\ \midrule
CoLink(Ours)            & 3536.07                                          & 4840.54                           & 0.0313            & 2836.89                      & 4843.31             & 0.0156            \\  \bottomrule
\end{tabular}
}
\end{table*}
\textbf{CoLink achieves superior compression efficiency.} Beyond generation quality, we evaluate the compression ratio of various approaches, as summarized in Table~\ref{tab:com-ratio}. CoLink reaches optimal performance utilizing only $3.1\%$ of the original context length. This evidence reaffirms the high efficiency and effectiveness of our cognitive-inspired cross-semantic linking and IB-driven abstraction.
Unlike traditional structured approaches, CoLink directly compresses the dialogue history in the latent space, circumventing costly entity extraction and graph construction to achieve $O(1)$ retrieval complexity. Furthermore, let $T$ denote the number of dialogue turns and $L$ the average length per turn, yielding a total historical context length of $N = T \cdot L$. Standard dense attention mechanisms suffer from quadratic time complexity $O((TL)^2)$ and linear space complexity $O(TL)$ due to the continuous growth of the KV cache. In contrast, CoLink compresses the unbounded history into $M$ fixed-size slots (where $M \ll TL$). This reduces the storage complexity for history representation to $O(M)$ and the attention complexity during generation to $O(M^2)$, ensuring constant-time scalability regardless of dialogue duration.

\begin{table}[t]
  \caption{Transfer performance of CoLink on Llama-3.2-3B, involving only first-stage memory slot training.}
  \label{tab:model-transfer}
  \resizebox{\linewidth}{!}{
  \begin{tabular}{lcccccccc}
    \toprule
    \multicolumn{1}{c}{\multirow{2}{*}{\textbf{Input Encode}}} & \multicolumn{4}{c}{\textbf{BLEU} (\%)} & \multicolumn{3}{c}{\textbf{ROUGE} (\%)} & \multirow{2}{*}{\textbf{BS} (\%)} \\ \cmidrule(lr){2-5} \cmidrule(lr){6-8}
    \multicolumn{1}{c}{} & \textbf{B1} & \textbf{B2} & \textbf{B3}        & \textbf{B4} & \textbf{R1} & \textbf{R2} & \textbf{RL} & \\ \midrule
    Base(Full History)    & 18.21 & 11.02 & 7.84 & 4.66 & 30.03 & 10.13 & 15.31 & 69.38 \\
    Base(Summary History) & 21.89 & 13.11 & 7.65 & \textbf{5.57} & \textbf{32.36} & \textbf{11.21} & 17.09 & \textbf{70.46} \\
    Memory Slot(Ours)     & \textbf{24.21} & \textbf{14.04} & \textbf{8.60} & 5.02 & 31.82 & 10.44 & \textbf{17.70} & 70.25 \\
    \bottomrule
  \end{tabular}
  }
\end{table}

\subsubsection{Generalizability of CoLink}\label{sec:ex-general}
To further investigate the robustness of our framework, we evaluate CoLink across three critical dimensions: architectural transferability, cross-domain adaptability, and cross-dataset generalizability.

\textbf{Architectural Transferability.} To assess whether CoLink is model-agnostic, we apply the framework to Llama-3.2-3B \cite{touvron2023llama} by training the memory slots on the MMD dataset. As presented in Table~\ref{tab:model-transfer}, CoLink yields consistent performance gains over the vanilla Llama-3.2 architecture. This result underscores its strong architectural transferability, indicating that our cognitive-inspired compression mechanism can effectively augment various LLM backbones.

\begin{table}[t]
  \caption{Transfer performance of CoLink on English open-domain multi-turn dialogue dataset.}
  \label{tab:domain-adaption}
  \resizebox{\linewidth}{!}{
  \begin{tabular}{lcccccccc}
    \toprule
    \multicolumn{1}{c}{\multirow{2}{*}{\textbf{Input Encode}}} & \multicolumn{4}{c}{\textbf{BLEU} (\%)} & \multicolumn{3}{c}{\textbf{ROUGE} (\%)} & \multirow{2}{*}{\textbf{BS} (\%)} \\ \cmidrule(lr){2-5} \cmidrule(lr){6-8}
    \multicolumn{1}{c}{} & \textbf{B1} & \textbf{B2} & \textbf{B3}        & \textbf{B4} & \textbf{R1} & \textbf{R2} & \textbf{RL} & \\ \midrule
    Base(Full History)      & 23.28                  & 12.67                  & 7.97           & 5.12                   & 28.72                  & 8.62                   & 25.60                  & 74.96                                          \\
Base(Summary History)   & 18.18                  & 9.78                   & 6.03           & 3.75                   & 30.38                  & 9.02                   & 27.03                  & 74.16                                          \\
Memory Slot(Ours)       & \textbf{38.49}         & \textbf{23.16}         & \textbf{15.59} & \textbf{10.48}         & \textbf{39.42}         & \textbf{14.52}         & \textbf{35.33}         & \textbf{78.68}  \\
    \bottomrule
  \end{tabular}
  }
\end{table}

\textbf{Cross-Domain Adaptability.} We further evaluate CoLink on UltraData \cite{ding2023enhancing}, an English open-domain multi-turn dialogue dataset, to verify its performance beyond the medical domain. Notably, since external knowledge is not natively required for this task, we deploy only the memory slots for history compression. As shown in Table~\ref{tab:domain-adaption}, CoLink outperforms the full-history concatenation baseline by $8.16\%$, demonstrating that its IB-driven abstraction maintains high efficacy even in knowledge-scarce, open-domain scenarios.

\textbf{Cross-Dataset Generalizability.} Lastly, we evaluate CoLink's inference capacity by directly transferring the MMD-trained model to the CMtMedQA dataset without further fine-tuning. This cross-dataset setting, summarized in Table~\ref{tab:method-gener}, reveals that CoLink consistently outperforms standard baselines. Such results highlight its robust capacity for in-domain transfer and its ability to handle distribution shifts between different clinical dialogue contexts.
 \begin{table}[t]
  \caption{Generalization across different datasets.}
  \label{tab:method-gener}
  \resizebox{\linewidth}{!}{
  \begin{tabular}{lcccccccc}
    \toprule
    \multicolumn{1}{c}{\multirow{2}{*}{\textbf{Method}}} & \multicolumn{4}{c}{\textbf{BLEU} (\%)} & \multicolumn{3}{c}{\textbf{ROUGE} (\%)} & \multirow{2}{*}{\textbf{BS} (\%)} \\ \cmidrule(lr){2-5} \cmidrule(lr){6-8}
    \multicolumn{1}{c}{} & \textbf{B1} & \textbf{B2} & \textbf{B3}        & \textbf{B4} & \textbf{R1} & \textbf{R2} & \textbf{RL} & \\ \midrule
    \multicolumn{1}{l}{CoLink (stage1)} & \multicolumn{1}{l}{26.40} & 13.40 & 7.88 & 4.54 & \multicolumn{1}{l}{29.09} & 8.63 & 20.52 & \multicolumn{1}{l}{73.30} \\
    \bottomrule
  \end{tabular}
  }
  \vskip -0.1in
\end{table}

\subsubsection{Effectiveness Analysis}\label{sec:ex-effect-an}
We theoretically analyze the superiority of CoLink over sequential summarization from both forward (attention) and backward (gradient) perspectives.

\textbf{Attention Competition.} To compress a long text of length $L$ into $M$ memory slots, the information that an input token $j$ contributes to a slot $i$ is formulated as:
\begin{equation}
A_{i,j} = \frac{\exp(q_i \cdot k_j)}{\sum_{l=1}^M \exp(q_l \cdot k_j)} \label{eq22}
\end{equation}

This indicates that the same input token $j$ must vote and allocate its weight across all $M$ slots, facilitating a competitive absorption of the contextual information. In contrast, transformer-based summarization typically relies on hidden states or generates a short text directly. The attention between a summary token $i$ and an input token $j$ is calculated as:

\begin{equation}
A_{i,j} = \frac{\exp(q_i \cdot k_j)}{\sum_{l=1}^{L}\exp(q_i \cdot k_l)} \label{eq23}
\end{equation}

Compared to the competitive allocation of slots, this standard aggregation acts merely as a weighted average over the $L$ input tokens. As the input sequence length $L$ grows, this Softmax distribution inevitably flattens, inducing severe attention dilution and feature blurring.

\textbf{Gradient Saliency.} Furthermore, we analyze the optimization dynamics from a backward propagation perspective. Let $C = [\text{slots}][\text{query}]$ and $C = [\text{summary}][\text{query}]$ denote the corresponding LLM input formats during inference.

The loss function $\mathcal{L}$ inherently depends on the attention matrix $A$ (across any layer and head). By applying a first-order Taylor expansion \cite{yin2025clearsight}, we obtain the loss perturbation:
\begin{equation}
    \begin{aligned} \mathcal{L}(A + \Delta A) \approx \mathcal{L}(A) + \left\langle \frac{\partial \mathcal{L}}{\partial A}, \Delta A \right\rangle \\ \Delta \mathcal{L} \approx \left\langle \frac{\partial \mathcal{L}}{\partial A}, A \right\rangle = \sum_{i,j} A_{i,j} \cdot \frac{\partial \mathcal{L}}{\partial A_{i,j}} \end{aligned} \label{eq24}
\end{equation}

Compared to the relatively sparse representations in standard summarization, CoLink encapsulates denser semantics, thereby contributing higher information saliency ($\Delta \mathcal{L}$) to the optimization objective. Specifically, in sequential summarization, all gradients from a monotonic summary are forced through a shared representation channel, inevitably causing semantic interference. Conversely, in our proposed method, since the matrix $A_{i,j}$ strictly routes tokens via the aforementioned competition mechanism, the loss gradients $\frac{\partial \mathcal{L}}{\partial A_{i,j}}$ can backpropagate through highly orthogonal and decoupled paths, fundamentally preserving distinct and fine-grained features.

\begin{figure}[]
    \centerline{\includegraphics[width=\columnwidth]{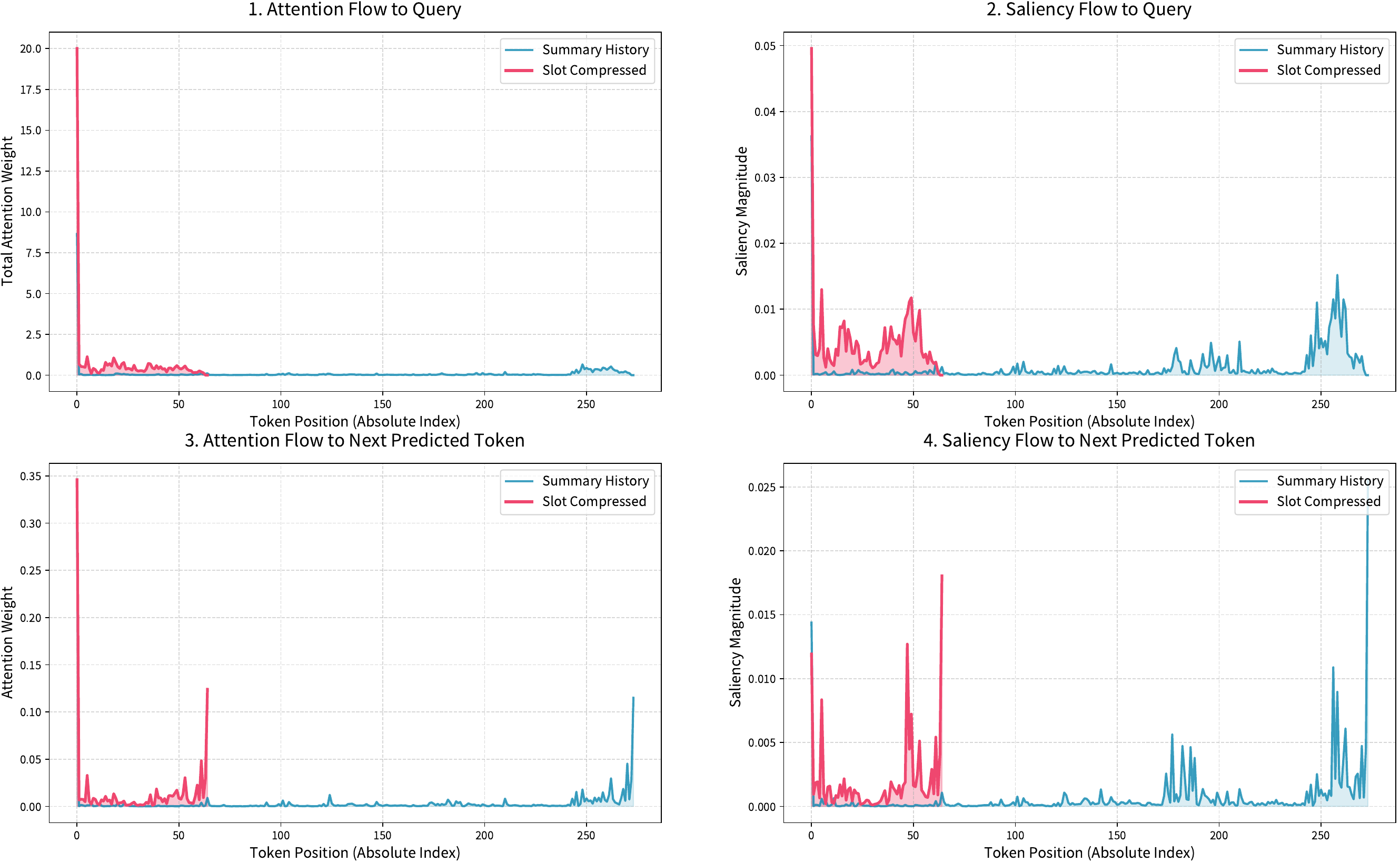}}
    \caption{
      Attention and loss saliency of summary history and memory slots towards the query and next-token prediction.
    }
    \label{fig:attn-saliency}
\end{figure}

We explicitly visualize this in Fig.~\ref{fig:attn-saliency}. The figure illustrates the attention and saliency distributions of a dialogue sample, comparing the standard summary history against memory slot compression. Excluding structural attention sinks (e.g., the initial token and the terminal query tokens), the compact slot representations (red lines) consistently exhibit significantly higher attention mass and gradient saliency towards both the query and the next-token prediction. This empirical observation perfectly corroborates our theoretical analysis: compared to the diluted and flattened distribution of the sequential summary (blue lines), CoLink inherently guarantees a highly condensed and salient feature space.

\subsubsection{Human and LLM-as-a-Judge Evaluation}\label{sec:ex-human-llm}
\begin{table*}[ht]
\centering
\caption{Human and LLM-as-a-Judge evaluation on Accuracy \& Professionalism, Informativeness, and Contextual Coherence.}
\label{tab:human-llm-evaluation}
\resizebox{\textwidth}{!}{%
\begin{tabular}{lcccccccc}
\toprule
\multicolumn{1}{c}{\multirow{2}{*}{\textbf{Method}}} & \multicolumn{4}{c}{\textbf{Human}}                                                                                                                  & \multicolumn{4}{c}{\textbf{LLM-as-Judge}}                                                                                                                               \\ \cmidrule(lr){2-5} \cmidrule(lr){6-9}
                        & \multicolumn{1}{c}{\textbf{Accuracy \& Professionalism}} & \multicolumn{1}{c}{\textbf{Informativeness}} & \multicolumn{1}{c}{\textbf{Contextual Coherence}} & \multicolumn{1}{c}{\textbf{Overall}} & \multicolumn{1}{c}{\textbf{Accuracy \& Professionalism}} & \multicolumn{1}{c}{\textbf{Informativeness}} & \multicolumn{1}{c}{\textbf{Contextual Coherence}} & \multicolumn{1}{c}{\textbf{Overall}} \\ \midrule
MemoryBank              & \textbf{3.77}               & \textbf{3.08}   & 4.28                 & 3.71          & \textbf{3.94}               & \textbf{3.49}   & 4.46                 & 3.96          \\
NV-embed                & 3.56                        & 2.72            & 4.14                 & 3.47          & 3.82                        & 3.24            & 4.31                 & 3.79          \\
GraphRAG                & 3.54                        & 2.44            & 4.65                 & 3.54          & 3.21                        & 2.56            & 4.42                 & 3.40          \\
CCA                     & 1.00                        & 1.00            & 1.00                 & 1.00          & 1.00                        & 1.00            & 1.00                 & 1.00          \\
MemTree                 & 2.80                        & 1.84            & 4.40                 & 3.01          & 3.78                        & 3.03            & 4.70                 & 3.84          \\
RAPTOR                  & 3.13                        & 1.91            & 3.98                 & 3.01          & 3.54                        & 2.64            & 4.38                 & 3.52          \\
HippoRAG2               & 3.03                        & 2.00            & 3.97                 & 3.00          & 3.14                        & 2.41            & 4.45                 & 3.33          \\
CoLink(Ours)            & \textbf{3.71}               & \textbf{3.00}   & \textbf{4.98}        & \textbf{3.90} & \textbf{3.90}               & \textbf{3.33}   & \textbf{4.94}        & \textbf{4.06} \\ \bottomrule
\end{tabular}
}
\end{table*}

We employ a dual-track evaluation strategy involving human experts and GPT-4o-mini \cite{achiam2023gpt} to rate responses on a 5-point scale across three specific dimensions: \textbf{Accuracy \& Professionalism}: Assesses whether the response is factually correct, free from hallucinations, and adheres to clinical standards. \textbf{Informativeness}: Measures the richness and utility of the content, penalizing vague or generic responses that fail to address specific queries. \textbf{Contextual Coherence}: Evaluates the model's ability to maintain logical consistency across turns and correctly reference historical details without contradiction.

Table~\ref{tab:human-llm-evaluation} summarizes the results. CoLink consistently achieves superior ratings across all metrics. Notably, it surpasses the strongest baseline by a substantial margin in Contextual Coherence, highlighting its robustness in modeling long-horizon dependencies.

\subsection{Ablation Studies}\label{sec:ex-ablation}
We design ablation studies to dissect the contributions of CoLink's components, including component analysis, history clustering strategies, loss sensitivity, and the impact of the input encoding space.
\begin{table*}[ht]
\centering
\caption{Performance after ablating memory slots and knowledge slots.}
\label{tab:ablation-compnent}
\resizebox{\textwidth}{!}{%
\begin{tabular}{ccccccccccccccccc}
\toprule
\multicolumn{1}{c}{\multirow{2}{*}{\textbf{History Len.}}} & \multicolumn{8}{c}{\textbf{MMD} (\%)}                                                                                                                                              & \multicolumn{8}{c}{\textbf{CMtMedQA} (\%)}                                                                                                                                             \\ \cmidrule(lr){2-9} \cmidrule(lr){10-17}
                        & \textbf{B1} & \textbf{B2} & \textbf{B3}        & \textbf{B4} & \textbf{R1} & \textbf{R2} & \textbf{RL} & \textbf{BS} & \textbf{B1}              & \textbf{B2}              & \textbf{B3}              & \textbf{B4}              & \textbf{R1}              & \textbf{R2}              & \textbf{RL}              & \textbf{BS}       \\ \midrule
CoLink(Ours)            & 27.52        & \textbf{13.87}        & 7.66 & \textbf{4.10}                 & 31.63        & \textbf{8.64}        & 19.26        & \textbf{72.84}                 & \textbf{31.62} & \textbf{18.02} & \textbf{11.74} & \textbf{7.62} & \textbf{34.28} & \textbf{12.31} & \textbf{25.50} & \textbf{75.36}           \\
w/o Memory Slots        & 24.13                 & 11.42                 & 6.26 & 3.35                 & 28.35                 & 7.08                 & 17.87                 & 70.94    & 28.73 & 15.83 & 9.96 & 6.18 & 31.89 & 10.81 & 23.74 & 74.33         \\
w/o Knowledge Slots     & \textbf{27.61}         & 13.82         & \textbf{7.71} & 3.96          & \textbf{31.88}         & 8.54          & \textbf{19.33}         & 72.78                               & 30.08  & 17.00  & 10.91  & 6.97   & 32.73           & 11.62  & 24.01           & 74.81    \\ \bottomrule
\end{tabular}
}
\end{table*}
\subsubsection{Ablation of Model Components}\label{sec:ex-component}
Table~\ref{tab:ablation-compnent} details the impact of removing the memory and knowledge slots. Ablating memory slots induces severe degradation. This strongly validates our IB-driven objective, which effectively distills critical state information from redundant history while filtering noise. It underscores that accurate context state maintenance is a prerequisite for reliable decision-making in multi-turn medical consultations. In contrast, the gains from knowledge activation and compression are relatively marginal. We attribute this to two factors: 1. \textit{Knowledge Misalignment}: Since neither MMD nor CMtMedQA possesses a natively aligned KG, we rely on general medical KGs. These external KGs often contain irrelevant entities and lack scenario-specific details (e.g., irrelevant diseases and lifestyle advice). This misalignment imposes an upper bound on the potential performance gain. 2. \textit{Semantic Sparsity}: Unlike dense textual history, KGs rely on discrete symbolic triples. This structural representation inherently suffers from semantic information loss, making it more challenging for the model to extract nuanced cues than from the rich natural language context.

To investigate this, we conduct experiments on the GMD dataset, which features a precisely aligned KG for symptom-based disease diagnosis \cite{yan2024eirad}. Table~\ref{tab:gmd} presents the diagnostic accuracy of various methods. \textit{Notably, the IB-driven knowledge slots yield a positive impact, demonstrating their effectiveness when accurate external knowledge is available.}

\begin{table}[t]
\centering
  \caption{Performance Comparison on the GMD Dataset with Precisely Aligned Knowledge Graphs.}
  \label{tab:gmd}
  \resizebox{0.8\linewidth}{!}{
  \begin{tabular}{lc}
    \toprule
    \textbf{Method}          & \textbf{Accuracy (\%)} \\ \midrule
Base                     & 79.08                  \\
+Full Knowledge            & 82.00                  \\
+Summary Knowledge         & 59.83                  \\
CoLink (Knowledge Slots) & \textbf{90.37}
  \\
    \bottomrule
  \end{tabular}
  }
\end{table}

\subsubsection{Sensitivity to Loss Configuration}\label{sec:ex-loss}
\begin{figure}[]
    \centerline{\includegraphics[width=\columnwidth]{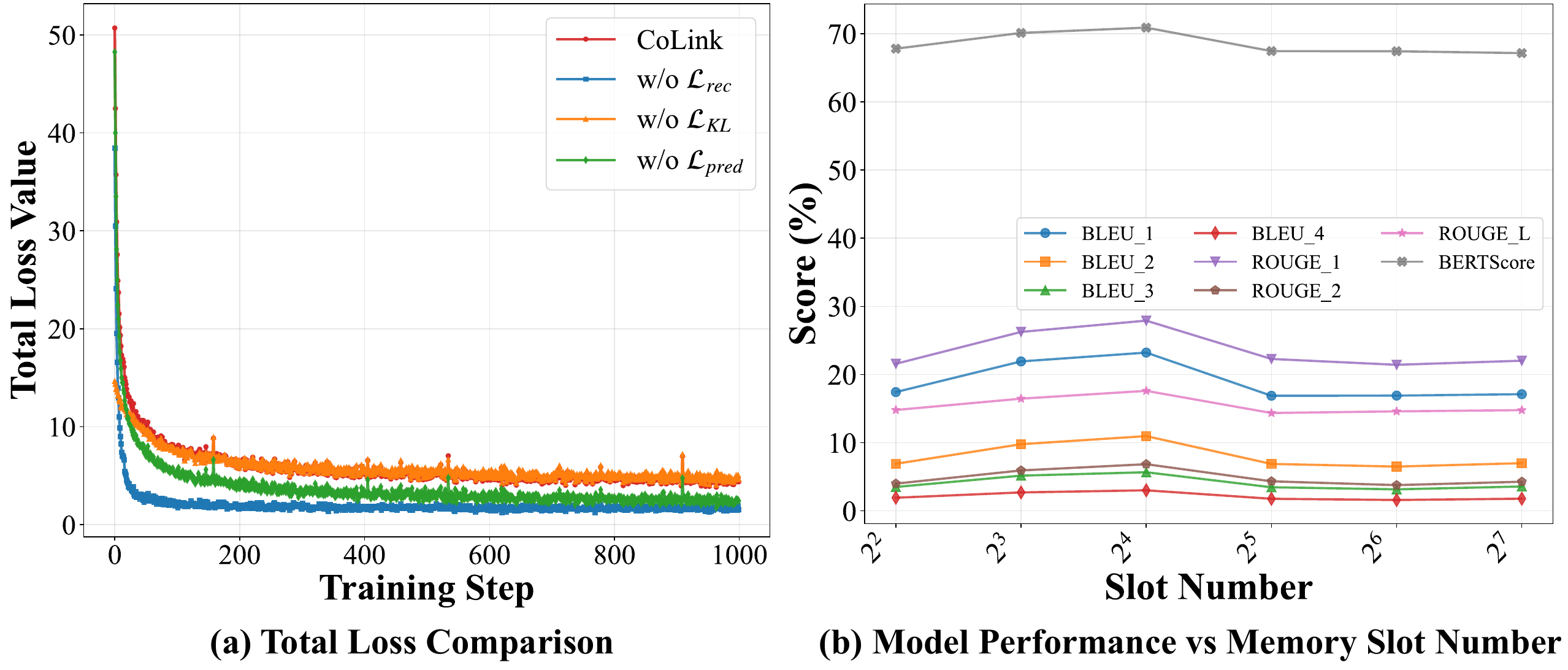}}
    \caption{
      Training loss trajectories under different ablation settings.
    }
    \label{fig:total-loss-and-slot}
\end{figure}

\begin{table}[t]
  \caption{Ablation study on different loss terms.}
  \label{tab:loss-ablation}
  \resizebox{0.98\linewidth}{!}{
  \begin{tabular}{lcccccccc}
    \toprule
    \multicolumn{1}{c}{\multirow{2}{*}{\textbf{Method}}} & \multicolumn{4}{c}{\textbf{BLEU} (\%)} & \multicolumn{3}{c}{\textbf{ROUGE} (\%)} & \multirow{2}{*}{\textbf{BS} (\%)} \\ \cmidrule(lr){2-5} \cmidrule(lr){6-8}
    \multicolumn{1}{c}{} & \textbf{B1} & \textbf{B2} & \textbf{B3}        & \textbf{B4} & \textbf{R1} & \textbf{R2} & \textbf{RL} & \\ \midrule
    CoLink (stage1)          & 23.21 & 10.96 & 5.66 & 3.01 & 27.90 & \textbf{6.84} & \textbf{17.60} & 70.88 \\
    w/o $\mathcal{L}_{rec}$  & 16.74 & 6.60 & 3.24 & 1.73 & 21.36 & 3.88 & 14.36 & 67.84 \\
    w/o $\mathcal{L}_{KL}$   & \textbf{25.65} & \textbf{11.81} & \textbf{5.99} & \textbf{3.15} & \textbf{29.27} & 6.81 & 17.33 & \textbf{71.36} \\
    w/o $\mathcal{L}_{pred}$ & 17.13 & 6.74 & 3.23 & 1.60 & 22.54 & 4.19 & 15.25 & 67.73 \\
    
    \bottomrule
  \end{tabular}
  }
\end{table}
To analyze the effects of the tripartite objective defined in Section~\ref{sec:me-obj}, we conduct loss ablations on the MMD dataset during first-stage memory slot training. The results are reported in Table~\ref{tab:loss-ablation}. Removing either $\mathcal{L}_{pred}$ or $\mathcal{L}_{rec}$ results in significant degradation, confirming their roles in preventing hallucination and semantic drift, respectively. Intriguingly, omitting the KL constraint ($\mathcal{L}_{KL}$) yields marginal metric improvements. We attribute this to the misalignment between the isotropic Gaussian prior and the pre-trained LLM's highly anisotropic embedding space \cite{ethayarajh2019contextual}. This suggests a trade-off between strict regularization (for OOD robustness, see Table~\ref{tab:method-gener}) and surface-level n-gram matching. We validate this trade-off through two auxiliary analyses: \textbf{Factor Scaling:} Varying the $\mathcal{L}_{KL}$ weight (Table~\ref{tab:scale-kl}) confirms that weaker regularization correlates with higher surface metrics. \textbf{Training Dynamics:} As shown in Fig.~\ref{fig:total-loss-and-slot}(a), removing $\mathcal{L}_{KL}$ leads to a slower convergence rate, indicating that although $\mathcal{L}_{KL}$ incurs a minor fitting penalty, it is essential for constraining representational dispersion and guiding stable convergence.
\begin{table}[t]
  \caption{The performance after scaling $\mathbb{L}_{KL}$.}
  \label{tab:scale-kl}
  \resizebox{\linewidth}{!}{
  \begin{tabular}{lcccccccc}
    \toprule
    \multicolumn{1}{c}{\multirow{2}{*}{\textbf{$\alpha$}}} & \multicolumn{4}{c}{\textbf{BLEU} (\%)} & \multicolumn{3}{c}{\textbf{ROUGE} (\%)} & \multirow{2}{*}{\textbf{BS} (\%)} \\ \cmidrule(lr){2-5} \cmidrule(lr){6-8}
    \multicolumn{1}{c}{} & \textbf{B1} & \textbf{B2} & \textbf{B3}        & \textbf{B4} & \textbf{R1} & \textbf{R2} & \textbf{RL} & \\ \midrule
    $10.0$ & 20.30 & 8.74 & 4.40 & 2.18 & 24.30 & 5.11 & 15.74 & 69.28 \\
    $0.1$  & \textbf{21.32} & \textbf{9.40} & \textbf{4.96} & \textbf{2.63} & \textbf{26.46} & \textbf{5.95} & \textbf{16.73} & \textbf{69.57} \\
    \bottomrule
  \end{tabular}
  }
  \vskip -0.1in
\end{table}

\subsubsection{Analysis of Input Encoding Space}\label{sec:ex-encoding}
\begin{table}[t]
  \caption{Impact of Input Encoding Methods on Model Performance}
  \label{tab:input-encode}
  \resizebox{0.98\linewidth}{!}{
  \begin{tabular}{lcccccccc}
    \toprule
    \multicolumn{1}{c}{\multirow{2}{*}{\textbf{Input Encode}}} & \multicolumn{4}{c}{\textbf{BLEU} (\%)} & \multicolumn{3}{c}{\textbf{ROUGE} (\%)} & \multirow{2}{*}{\textbf{BS} (\%)} \\ \cmidrule(lr){2-5} \cmidrule(lr){6-8}
    \multicolumn{1}{c}{} & \textbf{B1} & \textbf{B2} & \textbf{B3}        & \textbf{B4} & \textbf{R1} & \textbf{R2} & \textbf{RL} & \\ \midrule
    Layer 0 Hidden State    & 15.86 & 6.00 & 2.87 & 1.41 & 21.07 & 3.68 & 14.79 & 67.42 \\
    nn.embedding            & 17.14 & 6.74 & 3.21 & 1.60 & 21.64 & 3.67 & 14.43 & 67.93 \\
    Last Layer Hidden State & \textbf{23.21} & \textbf{10.96} & \textbf{5.66} & \textbf{3.01} & \textbf{27.90} & \textbf{6.84} & \textbf{17.60} & \textbf{70.88} \\
    \bottomrule
  \end{tabular}
  }
\end{table}

\begin{figure*}[h]
    \centerline{\includegraphics[width=\linewidth]{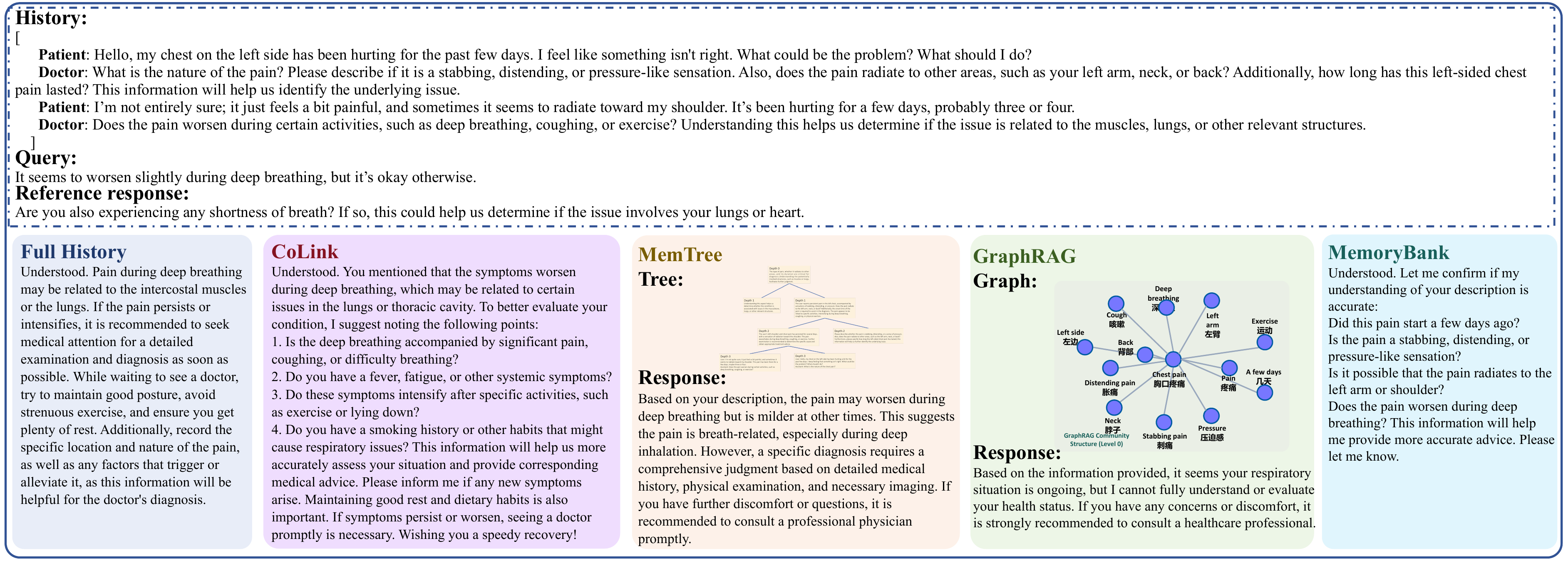}}
    \caption{
      Comparison of generations for a case across different models.
    }
    \label{fig:case-study}
\end{figure*}
Since CoLink encodes compressed slots as prefix prompts for the LLM, alignment between the compressor's semantic space and the LLM's native representation is critical. To mitigate semantic mismatch, our default implementation utilizes the LLM's own encoder to encode the context and extract the last hidden state for global abstraction. To quantify the impact of semantic drift introduced by different encoders, we compare this approach against two variants on dataset MMD: (1) a standard embedding layer (independent initialization), and (2) the LLM's first hidden layer. To strictly isolate the encoder's impact, we restrict these experiments to first-stage memory slot training and use a simplified input sequence: [Memory Slot] [Query]. Results are presented in Table~\ref{tab:input-encode}. We observe that maintaining alignment between the compression input and the LLM's internal representation yields noticeable performance gains. We hypothesize that deriving representations from the LLM's final layers ensures that the compressed slots reside within the same semantic manifold as the generative model. This alignment minimizes the adaptation gap, enabling the model to rapidly and accurately assimilate the core semantic signals encapsulated in the slots.

\subsubsection{Impact of History Clustering Strategies}\label{sec:ex-cluster}
\begin{table}[t]
  \caption{Performance comparison between compressing Retrieved Topic Clusters and Full History.}
  \label{tab:history-culster}
  \resizebox{\linewidth}{!}{
  \begin{tabular}{lcccccccc}
\toprule
\multicolumn{1}{c}{\multirow{2}{*}{\textbf{Method}}} & \multicolumn{4}{c}{\textbf{BLEU} (\%)} & \multicolumn{3}{c}{\textbf{ROUGE} (\%)} & \multirow{2}{*}{\textbf{BS} (\%)} \\ \cmidrule(lr){2-5} \cmidrule(lr){6-8}
    \multicolumn{1}{c}{} & \textbf{B1} & \textbf{B2} & \textbf{B3}        & \textbf{B4} & \textbf{R1} & \textbf{R2} & \textbf{RL} & \\ \midrule
History Cluster                             & 25.18          & 11.56          & \multicolumn{1}{l}{5.86} & 3.09          & \textbf{28.98} & 6.76          & \textbf{17.75} & 71.18                          \\
Full History                                & \textbf{25.58} & \textbf{11.86} & \textbf{6.15}            & \textbf{3.20} & 28.88          & \textbf{6.86} & 17.61          & \textbf{71.34}                 \\ \bottomrule
\end{tabular}
    }
  \vskip -0.1in
\end{table}
To evaluate the necessity of pre-retrieval clustering, we compare compressing retrieved topic clusters against compressing the full history. The comparative results for first-stage memory slot training are presented in Table~\ref{tab:history-culster}. We observe negligible performance variance between the two settings. We attribute this parity to two primary factors:
1. Robust Information Extraction: It demonstrates CoLink's intrinsic capability for noise filtration and semantic distillation. Even when exposed to the full unpruned history, the Information Bottleneck-guided compression effectively identifies and retains salient information while discarding redundancy.
2. High Information Density: Unlike sparse document retrieval tasks, domain-specific multi-turn dialogues (e.g., medical consultations) are characterized by high information density, where a larger portion of the history remains contextually relevant. This density likely diminishes the marginal gain of explicit clustering in this specific setting, though clustering remains computationally advantageous.

\subsection{Further Analysis}\label{sec:ex-further}
\subsubsection{Impact of Slot Number}\label{sec:ex-slot-num}
The number of memory slots ($M$) controls the granularity of global history compression. To investigate the trade-off between compression ratio and generation fidelity, we conduct sensitivity analysis by varying $M$ on MMD, as visualized in Fig.~\ref{fig:total-loss-and-slot}(b). Performance exhibits a non-monotonic trend: both overly small and excessively large $M$ lead to degradation. This behavior is theoretically intuitive. With too few slots, long dialogue histories are over-compressed, forcing each slot to encode excessive information and causing semantic loss. Conversely, too many slots dilute information density, leaving many slots underutilized or dominated by noise. This observation aligns with the \textit{local vs. global} trade-offs discussed in Section~\ref{sec:ex-main}.

\begin{table}[]
  \caption{AdaptSlot vs. Fixed-Slot.}
  \label{tab:adaptslot}
  \resizebox{\linewidth}{!}{
  \begin{tabular}{lcccccccc}
    \toprule
    \multicolumn{1}{c}{\multirow{2}{*}{\textbf{Method}}} & \multicolumn{4}{c}{\textbf{BLEU} (\%)} & \multicolumn{3}{c}{\textbf{ROUGE} (\%)} & \multirow{2}{*}{\textbf{BS} (\%)} \\ \cmidrule(lr){2-5} \cmidrule(lr){6-8}
    \multicolumn{1}{c}{} & \textbf{B1} & \textbf{B2} & \textbf{B3}        & \textbf{B4} & \textbf{R1} & \textbf{R2} & \textbf{RL} & \\ \midrule
    AdaptSlot (w/o Sparse Loss) & 6.65 & 2.51 & 1.43 & 0.88 & 15.45 & 2.36 & 10.72 & 60.53 \\
    AdaptSlot                   & 17.09 & 6.49 & 3.28 & 1.70 & 22.38 & 3.76 & 14.19 & 67.33 \\
    CoLink                      & \textbf{23.21} & \textbf{10.96} & \textbf{5.66} & \textbf{3.01} & \textbf{27.90} & \textbf{6.84} & \textbf{17.60} & \textbf{70.88} \\
    \bottomrule
  \end{tabular}
  }
  \vskip -0.1in
\end{table}
We further evaluate a dynamic-slot strategy adapted from AdaptSlot~\cite{fan2024adaptive}, which learns a masking function to prune redundant slots during training. The results are reported in Table~\ref{tab:adaptslot}. We find that dynamic slots often collapse into trivial behaviors: either masking the majority of slots or retaining the entire set, both of which substantially harm performance. Consequently, achieving stable, self-adaptive slot selection remains an open challenge.

\subsubsection{Case Study}\label{sec:ex-case}
Fig.~\ref{fig:case-study} presents a qualitative comparison of representative model outputs. MemTree offers a partially relevant response (e.g., the pain may be related to breathing), but it relies primarily on the immediate query and fails to incorporate valuable historical context. GraphRAG exhibits similar deficiencies. Visualizing their underlying structures reveals significant information loss, attributable to both extraction errors and the constraints of rigid structured representations (discussed in Section~\ref{sec:ex-main}). In contrast, CoLink demonstrates superior alignment with the Full History baseline. By leveraging soft memory slots, it avoids structural fragmentation and successfully synthesizes global context to deliver precise and actionable medical advice.

\section{Conclusion and Future Works}\label{sec:conclusion}
We revisit multi-turn dialogue through the lens of cognitive neuroscience and propose CoLink, a framework that innovatively integrates the IB principle of global abstraction with adaptive control of thought to activate associative knowledge. To address the scarcity of multi-turn dialogue data, we design a knowledge-augmented multi-agent debate framework and construct a dataset, MMD. Comprehensive evaluations demonstrate CoLink's significant superiority over strong baselines while utilizing only $3.125\%$ of the original context length. Driven by the IB principle, CoLink matches or even exceeds the performance of the full visible history. Furthermore, its soft, learnable slot mechanism preserves continuous semantic flow, avoiding the structural fragmentation inherent in rigid approaches. Rigorous validation through ablations and human evaluation confirms the robustness, model-agnostic nature, and generalization capability of our approach.

While theoretical derivations and extensive experiments validate CoLink's efficiency in history compression, our analysis reveals a trade-off: the method excels at capturing abstract semantic representations but struggles with tasks that require precise lexical retrieval. We evaluated this using the MemDaily dataset \cite{zhang2024memsim}, which features concise histories but demands high-precision recall of specific entities (e.g., phone numbers, job titles). On this benchmark, CoLink underperforms compared to text-based prompt baselines (such as Full History and Summary History). This suggests that the compression process may filter out fine-grained details necessary for exact factual reconstruction. Consequently, enhancing fidelity in precise information retrieval remains a pivotal direction for future work.


\bibliographystyle{IEEEtran}
\bibliography{ref.bib}

@article{chen2024core,
  title={Core context aware transformers for long context language modeling},
  author={Chen, Yaofo and You, Zeng and Zhang, Shuhai and Li, Haokun and Li, Yirui and Wang, Yaowei and Tan, Mingkui},
  journal={arXiv preprint arXiv:2412.12465},
  year={2024}
}

@article{lee2024nv,
  title={Nv-embed: Improved techniques for training llms as generalist embedding models},
  author={Lee, Chankyu and Roy, Rajarshi and Xu, Mengyao and Raiman, Jonathan and Shoeybi, Mohammad and Catanzaro, Bryan and Ping, Wei},
  journal={arXiv preprint arXiv:2405.17428},
  year={2024}
}

@article{edge2024local,
  title={From local to global: A graph rag approach to query-focused summarization},
  author={Edge, Darren and Trinh, Ha and Cheng, Newman and Bradley, Joshua and Chao, Alex and Mody, Apurva and Truitt, Steven and Metropolitansky, Dasha and Ness, Robert Osazuwa and Larson, Jonathan},
  journal={arXiv preprint arXiv:2404.16130},
  year={2024}
}

@article{rezazadeh2024isolated,
  title={From isolated conversations to hierarchical schemas: Dynamic tree memory representation for llms},
  author={Rezazadeh, Alireza and Li, Zichao and Wei, Wei and Bao, Yujia},
  journal={arXiv preprint arXiv:2410.14052},
  year={2024}
}

@article{gutierrez2025rag,
  title={From rag to memory: Non-parametric continual learning for large language models},
  author={Guti{\'e}rrez, Bernal Jim{\'e}nez and Shu, Yiheng and Qi, Weijian and Zhou, Sizhe and Su, Yu},
  journal={arXiv preprint arXiv:2502.14802},
  year={2025}
}

@article{jimenez2024hipporag,
  title={Hipporag: Neurobiologically inspired long-term memory for large language models},
  author={Jimenez Gutierrez, Bernal and Shu, Yiheng and Gu, Yu and Yasunaga, Michihiro and Su, Yu},
  journal={Advances in Neural Information Processing Systems},
  volume={37},
  pages={59532--59569},
  year={2024}
}

@article{wang2025archrag,
  title={ArchRAG: Attributed Community-based Hierarchical Retrieval-Augmented Generation},
  author={Wang, Shu and Fang, Yixiang and Zhou, Yingli and Liu, Xilin and Ma, Yuchi},
  journal={arXiv preprint arXiv:2502.09891},
  year={2025}
}

@article{xiao2023efficient,
  title={Efficient streaming language models with attention sinks},
  author={Xiao, Guangxuan and Tian, Yuandong and Chen, Beidi and Han, Song and Lewis, Mike},
  journal={arXiv preprint arXiv:2309.17453},
  year={2023}
}

@article{liu2024lost,
  title={Lost in the middle: How language models use long contexts},
  author={Liu, Nelson F and Lin, Kevin and Hewitt, John and Paranjape, Ashwin and Bevilacqua, Michele and Petroni, Fabio and Liang, Percy},
  journal={Transactions of the Association for Computational Linguistics},
  volume={12},
  pages={157--173},
  year={2024}
}

@article{li2024streamingdialogue,
  title={Streamingdialogue: Prolonged dialogue learning via long context compression with minimal losses},
  author={Li, Jia-Nan and Tu, Quan and Mao, Cunli and Yu, Zhengtao and Wen, Ji-Rong and Yan, Rui},
  journal={Advances in Neural Information Processing Systems},
  volume={37},
  pages={86074--86101},
  year={2024}
}

@article{achiam2023gpt,
  title={Gpt-4 technical report},
  author={Achiam, Josh and Adler, Steven and Agarwal, Sandhini and Ahmad, Lama and Akkaya, Ilge and Aleman, Florencia Leoni and Almeida, Diogo and Altenschmidt, Janko and Altman, Sam and Anadkat, Shyamal and others},
  journal={arXiv preprint arXiv:2303.08774},
  year={2023}
}

@article{wang2025learning,
  title={Learning to break: Knowledge-enhanced reasoning in multi-agent debate system},
  author={Wang, Haotian and Du, Xiyuan and Yu, Weijiang and Chen, Qianglong and Zhu, Kun and Chu, Zheng and Yan, Lian and Guan, Yi},
  journal={Neurocomputing},
  volume={618},
  pages={129063},
  year={2025},
  publisher={Elsevier}
}

@article{jiang2025knowledge,
  title={Knowledge assimilation: Implementing knowledge-guided agricultural large language model},
  author={Jiang, Jingchi and Yan, Lian and Liu, Haifeng and Xia, Zhenbo and Wang, Haotian and Yang, Yang and Guan, Yi},
  journal={Knowledge-based systems},
  volume={314},
  pages={113197},
  year={2025},
  publisher={Elsevier}
}

@inproceedings{zhong2024memorybank,
  title={Memorybank: Enhancing large language models with long-term memory},
  author={Zhong, Wanjun and Guo, Lianghong and Gao, Qiqi and Ye, He and Wang, Yanlin},
  booktitle={Proceedings of the AAAI Conference on Artificial Intelligence},
  volume={38},
  number={17},
  pages={19724--19731},
  year={2024}
}

@inproceedings{sarthi2024raptor,
  title={Raptor: Recursive abstractive processing for tree-organized retrieval},
  author={Sarthi, Parth and Abdullah, Salman and Tuli, Aditi and Khanna, Shubh and Goldie, Anna and Manning, Christopher D},
  booktitle={The Twelfth International Conference on Learning Representations},
  year={2024}
}

@inproceedings{cai2022memory,
  title={Memory graph with message rehearsal for multi-turn dialogue generation},
  author={Cai, Xiaoyu and Fu, Yao and Zhao, Hong and Jiang, Weihao and Pu, Shiliang},
  booktitle={Proceedings of the 31st ACM International Conference on Information \& Knowledge Management},
  pages={108--117},
  year={2022}
}

@article{huitt2003information,
  title={The information processing approach to cognition},
  author={Huitt, William},
  journal={Educational psychology interactive},
  volume={3},
  number={2},
  pages={53},
  year={2003}
}

@incollection{atkinson1968human,
  title={Human memory: A proposed system and its control processes},
  author={Atkinson, Richard C and Shiffrin, Richard M},
  booktitle={Psychology of learning and motivation},
  volume={2},
  pages={89--195},
  year={1968},
  publisher={Elsevier}
}

@article{yan2024eirad,
  title={EIRAD: An evidence-based dialogue system with highly interpretable reasoning path for automatic diagnosis},
  author={Yan, Lian and Guan, Yi and Wang, Haotian and Lin, Yi and Yang, Yang and Wang, Boran and Jiang, Jingchi},
  journal={IEEE Journal of Biomedical and Health Informatics},
  year={2024},
  publisher={IEEE}
}

@inproceedings{zhu2024information,
  title={An Information Bottleneck Perspective for Effective Noise Filtering on Retrieval-Augmented Generation},
  author={Zhu, Kun and Feng, Xiaocheng and Du, Xiyuan and Gu, Yuxuan and Yu, Weijiang and Wang, Haotian and Chen, Qianglong and Chu, Zheng and Chen, Jingchang and Qin, Bing},
  booktitle={Proceedings of the 62nd Annual Meeting of the Association for Computational Linguistics (Volume 1: Long Papers)},
  pages={1044--1069},
  year={2024}
}

@inproceedings{sun2024parrot,
  title={Parrot: Enhancing multi-turn instruction following for large language models},
  author={Sun, Yuchong and Liu, Che and Zhou, Kun and Huang, Jinwen and Song, Ruihua and Zhao, Wayne Xin and Zhang, Fuzheng and Zhang, Di and Gai, Kun},
  booktitle={Proceedings of the 62nd Annual Meeting of the Association for Computational Linguistics (Volume 1: Long Papers)},
  pages={9729--9750},
  year={2024}
}

@inproceedings{zhu2025knowledge,
  title={Knowledge Graph-Guided Retrieval Augmented Generation},
  author={Zhu, Xiangrong and Xie, Yuexiang and Liu, Yi and Li, Yaliang and Hu, Wei},
  booktitle={Proceedings of the 2025 Conference of the Nations of the Americas Chapter of the Association for Computational Linguistics: Human Language Technologies (Volume 1: Long Papers)},
  pages={8912--8924},
  year={2025}
}

@article{grootendorst2022bertopic,
  title={BERTopic: Neural topic modeling with a class-based TF-IDF procedure},
  author={Grootendorst, Maarten},
  journal={arXiv preprint arXiv:2203.05794},
  year={2022}
}

@article{tishby2000information,
  title={The information bottleneck method},
  author={Tishby, Naftali and Pereira, Fernando C and Bialek, William},
  journal={arXiv preprint physics/0004057},
  year={2000}
}

@article{locatello2020object,
  title={Object-centric learning with slot attention},
  author={Locatello, Francesco and Weissenborn, Dirk and Unterthiner, Thomas and Mahendran, Aravindh and Heigold, Georg and Uszkoreit, Jakob and Dosovitskiy, Alexey and Kipf, Thomas},
  journal={Advances in neural information processing systems},
  volume={33},
  pages={11525--11538},
  year={2020}
}

@inproceedings{fan2024adaptive,
  title={Adaptive slot attention: Object discovery with dynamic slot number},
  author={Fan, Ke and Bai, Zechen and Xiao, Tianjun and He, Tong and Horn, Max and Fu, Yanwei and Locatello, Francesco and Zhang, Zheng},
  booktitle={Proceedings of the IEEE/CVF Conference on Computer Vision and Pattern Recognition},
  pages={23062--23071},
  year={2024}
}

@article{ba2016layer,
  title={Layer normalization},
  author={Ba, Jimmy Lei and Kiros, Jamie Ryan and Hinton, Geoffrey E},
  journal={arXiv preprint arXiv:1607.06450},
  year={2016}
}

@inproceedings{dey2017gate,
  title={Gate-variants of gated recurrent unit (GRU) neural networks},
  author={Dey, Rahul and Salem, Fathi M},
  booktitle={2017 IEEE 60th international midwest symposium on circuits and systems (MWSCAS)},
  pages={1597--1600},
  year={2017},
  organization={IEEE}
}

@article{hu2022lora,
  title={Lora: Low-rank adaptation of large language models.},
  author={Hu, Edward J and Shen, Yelong and Wallis, Phillip and Allen-Zhu, Zeyuan and Li, Yuanzhi and Wang, Shean and Wang, Lu and Chen, Weizhu and others},
  journal={ICLR},
  volume={1},
  number={2},
  pages={3},
  year={2022}
}

@inproceedings{kong2024platolm,
  title={PlatoLM: Teaching LLMs in multi-round dialogue via a user simulator},
  author={Kong, Chuyi and Fan, Yaxin and Wan, Xiang and Jiang, Feng and Wang, Benyou},
  booktitle={Proceedings of the 62nd Annual Meeting of the Association for Computational Linguistics (Volume 1: Long Papers)},
  pages={7841--7863},
  year={2024}
}

@inproceedings{yang2024zhongjing,
  title={Zhongjing: Enhancing the chinese medical capabilities of large language model through expert feedback and real-world multi-turn dialogue},
  author={Yang, Songhua and Zhao, Hanjie and Zhu, Senbin and Zhou, Guangyu and Xu, Hongfei and Jia, Yuxiang and Zan, Hongying},
  booktitle={Proceedings of the AAAI conference on artificial intelligence},
  volume={38},
  number={17},
  pages={19368--19376},
  year={2024}
}

@article{bai2023qwen,
  title={Qwen technical report},
  author={Bai, Jinze and Bai, Shuai and Chu, Yunfei and Cui, Zeyu and Dang, Kai and Deng, Xiaodong and Fan, Yang and Ge, Wenbin and Han, Yu and Huang, Fei and others},
  journal={arXiv preprint arXiv:2309.16609},
  year={2023}
}

@article{li2025quality,
  title={Quality-Controllable automatic construction method of Chinese knowledge graph for medical decision-making applications},
  author={Li, Xue and Yuan, Ye and Yang, Yang and Guan, Yi and Wang, Haotian and Jiang, Jingchi and Shi, Huaizhang and Liu, Xiguang},
  journal={Information Processing \& Management},
  volume={62},
  number={4},
  pages={104148},
  year={2025},
  publisher={Elsevier}
}

@inproceedings{ethayarajh2019contextual,
  title={How Contextual are Contextualized Word Representations? Comparing the Geometry of BERT, ELMo, and GPT-2 Embeddings},
  author={Ethayarajh, Kawin},
  booktitle={Proceedings of the 2019 Conference on Empirical Methods in Natural Language Processing and the 9th International Joint Conference on Natural Language Processing (EMNLP-IJCNLP)},
  pages={55--65},
  year={2019}
}

@article{touvron2023llama,
  title={Llama: Open and efficient foundation language models},
  author={Touvron, Hugo and Lavril, Thibaut and Izacard, Gautier and Martinet, Xavier and Lachaux, Marie-Anne and Lacroix, Timoth{\'e}e and Rozi{\`e}re, Baptiste and Goyal, Naman and Hambro, Eric and Azhar, Faisal and others},
  journal={arXiv preprint arXiv:2302.13971},
  year={2023}
}

@article{byambasuren2019preliminary,
  title={Preliminary study on the construction of Chinese medical knowledge graph},
  author={Byambasuren, Odma and Yang, Yunfei and Sui, Zhifang and Dai, Damai and Chang, Baobao and Li, Sujian and Zan, Hongying},
  journal={Journal of Chinese Information Processing},
  volume={33},
  number={10},
  pages={1--9},
  year={2019}
}

@article{zhang2024memsim,
  title={Memsim: A bayesian simulator for evaluating memory of llm-based personal assistants},
  author={Zhang, Zeyu and Dai, Quanyu and Chen, Luyu and Jiang, Zeren and Li, Rui and Zhu, Jieming and Chen, Xu and Xie, Yi and Dong, Zhenhua and Wen, Ji-Rong},
  journal={arXiv preprint arXiv:2409.20163},
  year={2024}
}

@article{squire2004memory,
  title={Memory systems of the brain: a brief history and current perspective},
  author={Squire, Larry R},
  journal={Neurobiology of learning and memory},
  volume={82},
  number={3},
  pages={171--177},
  year={2004},
  publisher={Elsevier}
}

@article{popham2021visual,
  title={Visual and linguistic semantic representations are aligned at the border of human visual cortex},
  author={Popham, Sara F and Huth, Alexander G and Bilenko, Natalia Y and Deniz, Fatma and Gao, James S and Nunez-Elizalde, Anwar O and Gallant, Jack L},
  journal={Nature neuroscience},
  volume={24},
  number={11},
  pages={1628--1636},
  year={2021},
  publisher={Nature Publishing Group US New York}
}

@inproceedings{ding2023enhancing,
  title={Enhancing chat language models by scaling high-quality instructional conversations},
  author={Ding, Ning and Chen, Yulin and Xu, Bokai and Qin, Yujia and Hu, Shengding and Liu, Zhiyuan and Sun, Maosong and Zhou, Bowen},
  booktitle={Proceedings of the 2023 Conference on Empirical Methods in Natural Language Processing},
  pages={3029--3051},
  year={2023}
}

@article{sun2024massive,
  title={Massive activations in large language models},
  author={Sun, Mingjie and Chen, Xinlei and Kolter, J Zico and Liu, Zhuang},
  journal={First Conference on Language Modeling (COLM)},
  year={2024}
}

@inproceedings{zhang2024tell,
  title={Tell your model where to attend: Post-hoc attention steering for llms},
  author={Zhang, Qingru and Singh, Chandan and Liu, Liyuan and Liu, Xiaodong and Yu, Bin and Gao, Jianfeng and Zhao, Tuo},
  booktitle={International Conference on Learning Representations},
  volume={2024},
  pages={42411--42430},
  year={2024}
}

@article{gan2026unleashing,
  title={Unleashing diffusion transformers for visual correspondence by modulating massive activations},
  author={Gan, Chaofan and Tu, Yuanpeng and Chen, Xi and Chen, Tieyuan and Li, Yuxi and Harandi, Mehrtash and Lin, Weiyao},
  journal={Advances in Neural Information Processing Systems},
  volume={38},
  pages={114432--114462},
  year={2026}
}

@inproceedings{kang2025see,
  title={SEE WHAT YOU ARE TOLD: VISUAL ATTENTION SINK IN LARGE MULTIMODAL MODELS},
  author={Kang, Seil and Kim, Jinyeong and Kim, Junhyeok and Hwang, Seong Jae},
  booktitle={13th International Conference on Learning Representations, ICLR 2025},
  pages={13547--13574},
  year={2025},
  organization={International Conference on Learning Representations, ICLR}
}

@article{zhang2023h2o,
  title={H2o: Heavy-hitter oracle for efficient generative inference of large language models},
  author={Zhang, Zhenyu and Sheng, Ying and Zhou, Tianyi and Chen, Tianlong and Zheng, Lianmin and Cai, Ruisi and Song, Zhao and Tian, Yuandong and R{\'e}, Christopher and Barrett, Clark and others},
  journal={Advances in Neural Information Processing Systems},
  volume={36},
  pages={34661--34710},
  year={2023}
}

@article{liang2025ai,
  title={Ai meets brain: Memory systems from cognitive neuroscience to autonomous agents},
  author={Liang, Jiafeng and Li, Hao and Li, Chang and Zhou, Jiaqi and Jiang, Shixin and Wang, Zekun and Ji, Changkai and Zhu, Zhihao and Liu, Runxuan and Ren, Tao and others},
  journal={arXiv preprint arXiv:2512.23343},
  year={2025}
}

@article{liu2025efficient,
  title={Efficient Algorithms for Computing Random Walk Centrality},
  author={Liu, Changan and Xie, Zixuan and Zehmakan, Ahad N and Zhang, Zhongzhi},
  journal={IEEE Transactions on Knowledge and Data Engineering},
  year={2025},
  publisher={IEEE}
}

@inproceedings{yin2025clearsight,
  title={Clearsight: Visual signal enhancement for object hallucination mitigation in multimodal large language models},
  author={Yin, Hao and Si, Guangzong and Wang, Zilei},
  booktitle={Proceedings of the Computer Vision and Pattern Recognition Conference},
  pages={14625--14634},
  year={2025}
}

@article{qiang2026exploring,
  title={Exploring Knowledge Filtering for Retrieval-Augmented Question Answering},
  author={Qiang, Minjie and Wang, Zhongqing and Li, Shoushan and Zhou, Guodong},
  journal={IEEE Transactions on Audio, Speech and Language Processing},
  year={2026},
  publisher={IEEE}
}

\end{document}